\documentclass{article}[10]

\usepackage{microtype}
\usepackage{graphicx}
\usepackage{subfigure}
\usepackage{booktabs}

\usepackage[T1]{fontenc}

\usepackage{hyperref}

\usepackage[accepted]{icml2024}

\usepackage{amsmath}
\usepackage{amssymb}
\usepackage{mathtools}
\usepackage{amsthm}
\usepackage{xspace}
\usepackage{tcolorbox}
\usepackage{enumitem}
\setlist[enumerate]{topsep=0pt, partopsep=0pt, parsep=0pt, itemsep=0pt}
\setlist[itemize]{topsep=0pt, partopsep=0pt, parsep=0pt, itemsep=0pt}
\usepackage[skip=7pt]{caption}

\usepackage[capitalize,noabbrev]{cleveref}
\usepackage{xcolor}
\usepackage{colortbl}
\usepackage{soul}
\usepackage{adjustbox}
\usepackage{xurl}
\theoremstyle{plain}

\theoremstyle{definition}

\theoremstyle{remark}

\usepackage[textsize=tiny]{todonotes}

\begin{document}

\newcommand\revision[1]{\textcolor{black}{#1}}

\newcommand{\draftonly}[1]{#1} %
\newcommand{\draftcomment}[3]{\draftonly{{\color{#2}{{{[#1: #3]}}}}}}
\newcommand{\danqi}[1]{\draftcomment{Danqi}{orange}{#1}}
\newcommand{\alexis}[1]{\draftcomment{Alexis}{teal}{#1}}
\newcommand{\jiayi}[1]{\draftcomment{Jiayi}{green}{#1}}
\newcommand{\alex}[1]{\draftcomment{Alex}{blue}{#1}}
\newcommand{\simon}[1]{\draftcomment{Simon}{brown}{#1}}
\newcommand{\howard}[1]{\draftcomment{Howard}{magenta}{#1}}
\newcommand{\sebastian}[1]{\draftcomment{Sebastian}{yellow}{#1}}
\newcommand{\xindi}[1]{{\leavevmode\color[rgb]{1,0,1}[xindi: #1]}}

\newcommand{\tutoreval}{\textsc{TutorEval}\xspace}
\newcommand{\dataset}{\textsc{TutorChat}\xspace}
\newcommand{\mathdata}{MathMix\xspace}
\newcommand{\modelsmall}{Llemma-7B-32K-MathMix\xspace}
\newcommand{\modellarge}{Llemma-34B-MathMix\xspace}
\newcommand{\model}{Llemma-MathMix\xspace}

\definecolor{inputcolor}{HTML}{C9E4E7}
\newcommand{\hlinputcolor}[1]{\sethlcolor{inputcolor}\hl{#1}}
\definecolor{modeloutputcolor}{HTML}{FFE7B5}
\newcommand{\hlmodeloutputcolor}[1]{\sethlcolor{modeloutputcolor}\hl{#1}}
\definecolor{annotationcolor}{HTML}{99E5B5}
\newcommand{\hlannotationcolor}[1]{\sethlcolor{annotationcolor}\hl{#1}}

\twocolumn[
\icmltitle{Language Models as Science Tutors}
\icmltitlerunning{Language Models as Science Tutors}

\icmlsetsymbol{equal}{*}

\begin{icmlauthorlist}
\icmlauthor{Alexis Chevalier}{pli}
\icmlauthor{Jiayi Geng}{pli}
\icmlauthor{Alexander Wettig}{pli}
\icmlauthor{Howard Chen}{pli}
\icmlauthor{Sebastian Mizera}{iassns} \\
\vspace{0.5em}
\icmlauthor{Toni Annala}{iasmath}
\icmlauthor{Max Jameson Aragon}{neuro}
\icmlauthor{Arturo Rodr\'iguez Fanlo}{huji}
\icmlauthor{Simon Frieder}{ox}
\icmlauthor{Simon Machado}{eth}
\icmlauthor{Akshara Prabhakar}{pli}
\icmlauthor{Ellie Thieu}{wisc,meta}
\icmlauthor{Jiachen T. Wang}{pli}
\icmlauthor{Zirui Wang}{pli}
\icmlauthor{Xindi Wu}{pli}
\icmlauthor{Mengzhou Xia}{pli}
\icmlauthor{Wenhan Xia}{pli}
\icmlauthor{Jiatong Yu}{pli}
\icmlauthor{Jun-Jie Zhu}{env,and}
\icmlauthor{ Zhiyong Jason Ren}{env,and} \\
\vspace{0.5em}
\icmlauthor{Sanjeev Arora}{pli}
\icmlauthor{Danqi Chen}{pli}
\end{icmlauthorlist}

\icmlaffiliation{pli}{Princeton Language and Intelligence, Princeton University}
\icmlaffiliation{neuro}{Neuroscience Institute, Princeton University}
\icmlaffiliation{iassns}{School of Natural Sciences, Institute for Advanced Study}
\icmlaffiliation{iasmath}{School of Mathematics, Institute for Advanced Study}
\icmlaffiliation{huji}{Hebrew University of Jerusalem}
\icmlaffiliation{ox}{Department of Computer Science, Oxford University}
\icmlaffiliation{eth}{FIM - Institute for Mathematical Research, ETH Z\"urich}
\icmlaffiliation{wisc}{Department of Computer Science, University of Wisconsin-Madison}
\icmlaffiliation{meta}{Meta FAIR}
\icmlaffiliation{env}{Department of Civil and Environmental Engineering, Princeton University}
\icmlaffiliation{and}{Andlinger Center for Energy and the Environment, Princeton University}

\icmlcorrespondingauthor{Alexis Chevalier}{achevalier@princeton.edu}

\icmlkeywords{Machine Learning, ICML}

\vskip 0.12in
]

\printAffiliationsAndNotice{}  %

\begin{abstract}
NLP has recently made exciting progress toward training language models (LMs) with strong scientific problem-solving skills.
However, model development has not focused on real-life use-cases of LMs for science, including applications in education that require processing long scientific documents. To address this, we introduce  \tutoreval  and \dataset. \tutoreval is  a diverse question-answering benchmark consisting of  questions about long chapters from STEM textbooks, written by experts. \tutoreval helps measure real-life usability of LMs as scientific assistants, and it is the first benchmark combining long contexts, free-form generation, and multi-disciplinary scientific knowledge. Moreover, we show that fine-tuning base models with existing dialogue datasets leads to poor performance on \tutoreval. Therefore, we create \dataset, a dataset of 80,000 long synthetic dialogues about textbooks. We use \dataset to fine-tune Llemma models with 7B and 34B parameters. These LM tutors specialized in math have a 32K-token context window, and they excel at \tutoreval while performing strongly on GSM8K and MATH. Our datasets build on open-source materials, and we release our models, data, and evaluations publicly. 
\end{abstract}

\begin{figure}[!thb]
    \centering
    \begin{adjustbox}{center}
    \includegraphics[width=1.1\columnwidth]{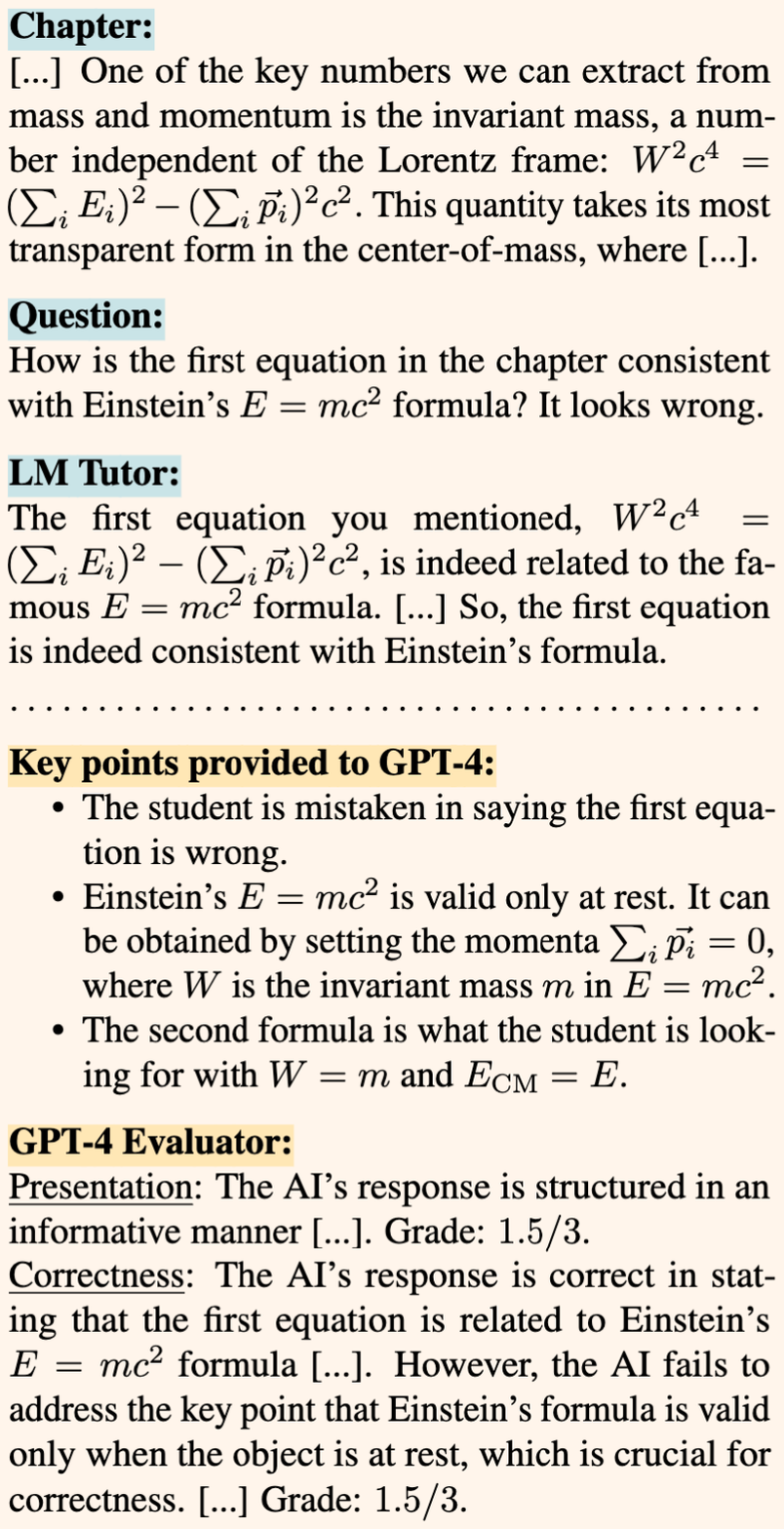}
    \end{adjustbox}
    \caption{
    Example from \tutoreval. Given the chapter, the student asks a question to the LM Tutor. Both the chapter and the question are fed to the LM Tutor to generate the answer. GPT-4 assesses the generation by referencing the human annotated key points (\hlinputcolor{blue}: the tutoring task; \hlmodeloutputcolor{yellow}: evaluation). See detailed examples in \S\ref{appdenix:eval_examples}.
    }
    \label{fig:teaser}
\end{figure}

\begin{figure*}[t]
    \centering
    \includegraphics[width=0.8\textwidth]{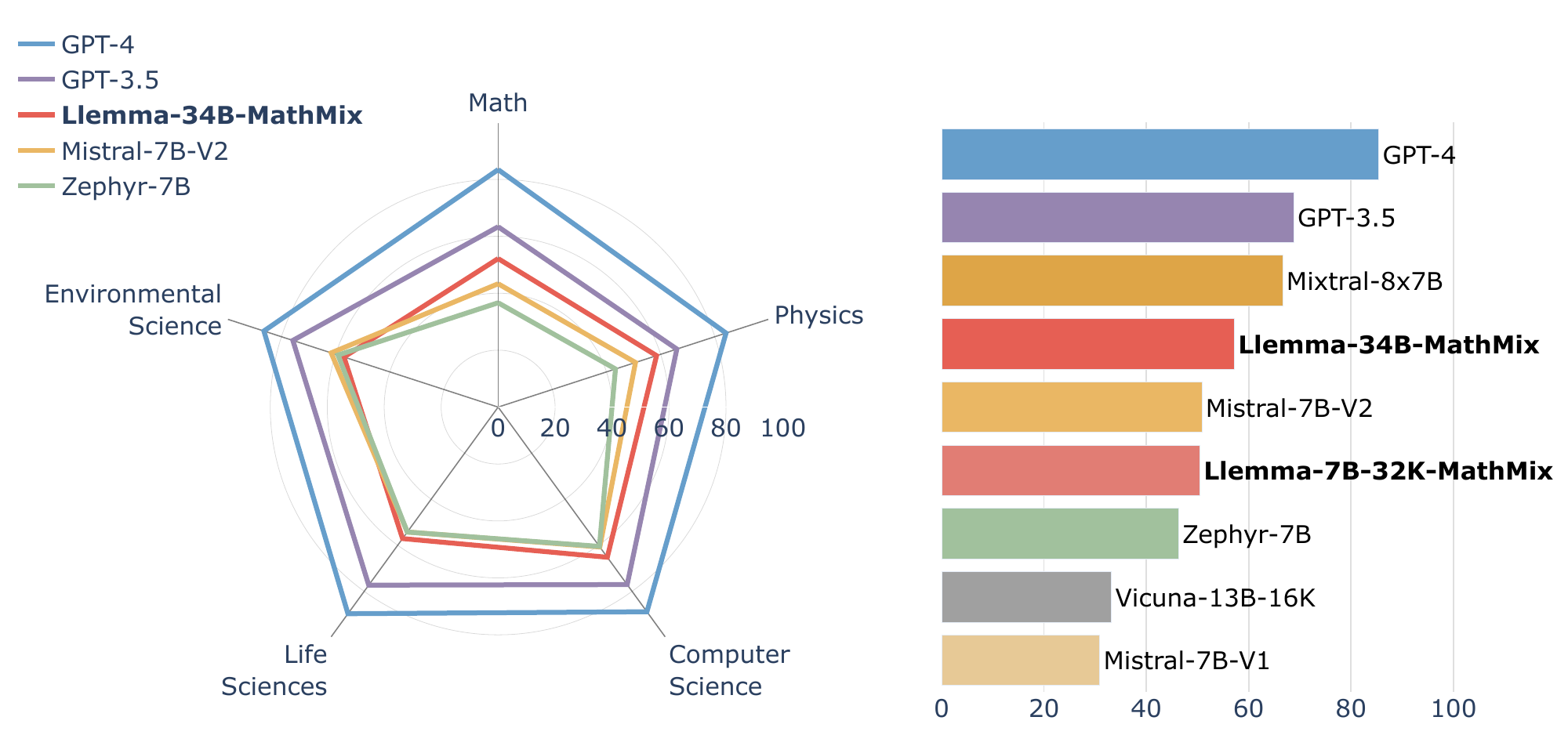}
    \caption{
        Left: performance breakdown on \tutoreval by domains. Right: leaderboard of popular models on \tutoreval. Our models, marked in bold, achieve competitive \tutoreval performance.
    }
    \label{fig:main_radar}
\end{figure*}
\section{Introduction}

NLP has recently made exciting progress towards improving LM performance on problem-solving benchmarks such as GSM8K \cite{gsm8k} or MATH \cite{hendrycksmath} and on general knowledge benchmarks like MMLU \cite{mmlu}.
One approach to constructing models for problem-solving consists in fine-tuning base models on math  datasets, such as MetaMath \cite{metamath}, MAmmoTH \cite{mammoth}, or WizardMath \cite{wizardmath}. Other approaches train on web-scraped corpora, such as Llemma \cite{llemma} with OpenWebMath \cite{openwebmath}, or on large synthetic datasets, such as Phi-2 \cite{textbooks1, textbooks2}.
This flurry of new LMs raises the question: how can we apply these problem-solving skills to build helpful LM tutors for science?  The first challenge for answering this question is the lack of relevant LM evaluations.

Existing science evaluations are lacking in the following ways. First, they do not simulate real-life situations where we use LMs as assistants, such as asking to clarify hard concepts or asking for background information. Second, science evaluations only evaluate the final answer to a problem, not the reasoning process. Third, real-life situations often require LMs to process long scientific materials, which are not tackled by any existing evaluations.

To bridge these gaps, we introduce \tutoreval, a long-context question-answering benchmark requiring advanced scientific knowledge, simulating humans seeking to understand textbook materials.  \tutoreval consists of over 800 questions written by experts in math, physics, computer science, environmental science, and life sciences. 
\tutoreval  extends the LM-as-evaluator framework \cite{alpaca_eval}  by introducing \emph{key points}: we collect sketches of the ground-truth answer from the experts, which an   LM evaluator uses to grade the LM tutor's answers. See Figure \ref{fig:teaser} for an example. We will show that GPT-4 can be used as a reliable evaluator for \tutoreval. 

We investigate the importance of fine-tuning for improving \tutoreval performance. We find that \tutoreval benefits from training on scientific texts and that fine-tuning with existing dialogue datasets delivers poor performance. Therefore, we create \dataset, the first long-context dialogue dataset for science. \dataset consists of 80,000  synthetic conversations about textbook chapters generated by GPT-3.5 and GPT-4, covering STEM topics, humanities and social sciences. 
We use \dataset  to fine-tune long-context models that rival strong closed-source models like Mistral-7B-V2 \cite{mistral} on \tutoreval.

We show that \dataset is a rich resource for domain-specific fine-tuning. 
While we find that it is tricky to acquire math skills during fine-tuning,  the  specificity of \dataset provides a remedy: combining STEM dialogues with math data yields well-rounded LM tutors with strong math problem-solving skills. With this insight, we construct a new data mixture, \mathdata, and we introduce \modelsmall and \modellarge, two long-context LMs competitive on \tutoreval, MATH and GSM8K.

In summary, we introduce \tutoreval, the first long-context science benchmark, and \dataset, the first long-context dialogue dataset for science,  and we conduct thorough investigations into the importance of training and fine-tuning with scientific texts. We release competitive long-context models specialized in science and math reasoning,  as well as all  our data and evaluations at \url{https://github.com/princeton-nlp/LM-Science-Tutor}.

\begin{table*}[ht]
\caption{\tutoreval question counts per domain and category. Question categories are described in Section \ref{sub:dataset_construction}.}
\centering

\begin{tabular}{lcccccc}
\toprule
                    & Math & CS & Physics & Env. & Life & Total \\
\midrule
Hard                & 96          & 59            & 54      & 22                 & 10        & 241   \\
Misleading          & 55          & 21            & 28      & 4                  & 5         & 113   \\
Closed-book         & 159         & 100           & 71      & 20                 & 20        & 370   \\
Answer in chapter   & 128         & 98            & 56      & 20                 & 6         & 308   \\
Total               & 362         & 205           & 174     & 53                 & 40        & 834   \\
\bottomrule
\end{tabular}

\label{tab:eval_composition}
\end{table*}

\section{Related Work}

\paragraph{LMs for science} Minerva \cite{minerva} and Galactica \cite{galactica} were two pioneering efforts to train LMs on large corpora of scientific documents to improve problem-solving and scientific knowledge.  Llemma \cite{llemma} has replicated  Minerva and Galactica's approach using the  open-source OpenWebMath dataset \cite{openwebmath}. MMLU \cite{mmlu} and MATH \cite{hendrycksmath} are key benchmarks which have motivated this general research direction. Other problem-solving benchmarks include SciBench \cite{scibench} and SciEval \cite{scieval}.

There has  been recent interest in developing the applications of LMs to education. EduBot \cite{edubot} applies LMs to English-teaching and MathDial \cite{mathdial} studies challenges of using LMs for math education, for which they release a small dataset of human-LM conversations. Various works have studied the expert-level capabilities of LMs and their applicability as teaching or research assistants. \citet{researchagents} benchmarks GPT-4 on machine learning research tasks. \citet{edqa_retrieval} studies retrieval-augmented capabilities of ChatGPT for education. General studies include  \citet{peskoff-stewart-2023-credible}, \citet{frieder2023mathematical}, and \citet{evaluating}.

\paragraph{Fine-tuning on model-generated dialogue} A popular approach for improving the instruction-following and dialogue capabilities of LMs consists in imitation learning: LMs are trained on the high-quality outputs of larger LMs. Examples include the textbooks of \citet{textbooks1} or Orca \cite{orca}.
The success of this distillation approach has led to an accumulation of  dialogue datasets. ShareGPT and WildChat \cite{wildchat} are efforts to crowd-source real-life ChatGPT conversations. UltraChat \cite{ultrachat} and OpenOrca \cite{openorca} are datasets of synthetic text generated by ChatGPT.

\paragraph{LM as an evaluator} The research community has recently explored new ways of evaluating  LMs. One solution resorts to using strong pre-trained LMs to compare model outputs. AlpacaFarm \cite{alpaca_eval} uses LMs to simulate human evaluators by obtaining preference votes between model outputs. 
\citet{llmhumaneval}, \citet{judgingllm}, \citet{chen2023exploring} analyse GPT-4's ability to simulate human preferences. LLMBar \cite{llmbar} introduces a systematic evaluation of LLMs as evaluators for instruction-following. Most works find that GPT-4 is generally reliable, and we build on these findings to introduce \tutoreval.

\section{\tutoreval}

In Section \ref{sub:eval_construction}, we show how \tutoreval was constructed with the help of experts and how \tutoreval can be used for both open-book and closed-book evaluation. In Section \ref{sub:llm_evaluator}, we explain how we use an LM as an evaluator and we show that GPT-4 correlates well with human judgments. In Section \ref{sub:leaderboard}, we comment on the leaderboard in Figure \ref{fig:main_radar}.
\subsection{Dataset construction}\label{sub:eval_construction}

\paragraph{Question collection} We work with 17 annotators, all STEM researchers with teaching experience. Each annotator was provided a set of textbook chapters 
and was instructed to write questions about chapters which they would be qualified to teach in a class. 
These chapters come from open-source textbooks collected from libretexts.org, as  part of the dataset discussed in Section \ref{sub:dataset_construction}.
Annotators were given free rein to write questions. They were only instructed to simulate questions that students could plausibly ask about each chapter. See   examples in Figure \ref{fig:teaser} and in Appendix \ref{appdenix:eval_examples}.

\paragraph{Dataset composition} In total, we collect 834 questions spanning math, physics,  computer science, life sciences and environmental science.  202 textbook chapters were used, with at least 3 questions per chapter. These chapters contain 1,800 words on average, and up to 6,100 words.  
We note that \tutoreval is biased toward math and computer science, but we hope to extend  \tutoreval in the future as follow-up work. See Table \ref{tab:eval_composition} for statistics.

\paragraph{Data categories} \tutoreval is a diverse set of questions allowing us to evaluate models in different settings and to track fine-grained behavior.  For each question, we report the following information: 
\begin{enumerate}[itemsep=0pt, parsep=0pt, topsep=0pt]
    \item \textit{Closed-book}: can the \textit{question} be understood without reading the chapter first?
    \item \textit{Answer in chapter}: is the \textit{answer} contained in the chapter, or does it require substantial external knowledge?
    \item \textit{Misleading}: is the question phrased in a misleading way, such as including false assumptions or an having overly-confident tone?
    \item \textit{Difficulty}: how hard is the question relatively to the chapter's intended audience?
\end{enumerate} 

\tutoreval contains 370 closed-book questions, meaning that the question does not refer explicitly to the chapter. This enables evaluation of short-context LMs, and we refer to this  evaluation setting as \tutoreval-\textsc{ClosedBook}.

\paragraph{Data validation}
To ensure the relevance of our data categories, we evaluate GPT-4 on different data categories both in open-book \tutoreval and in the closed-book setting (Table~\ref{tab:analysis_by_fields}).
First, we see that GPT-4's results on open-book questions are significantly affected in closed-book evaluation, proving that open-book questions require the chapter. 
We also see that closed-book questions slightly benefit from open-book evaluation and  that splitting questions by \textit{answer in chapter} shows how well GPT-4 is able to utilize long contexts. This demonstrates that \tutoreval is a rich benchmark which offers a thorough investigation of long-context understanding in the science domain.

\begin{table*}[ht]
\caption{\tutoreval scores of 4  models graded by humans and GPT-4. Models are introduced in Section \ref{section:models}. See Appendix \ref{appendix:detail_results}, Table \ref{tab:expanded_human_eval} for detailed human evaluation results and Appendix \ref{appendix:grading_analysis} for data analysis about GPT-4/human agreement.}
\centering
\begin{tabular}{lcccc}
\toprule
& \multicolumn{2}{c}{\textbf{Human \tutoreval}}   & \multicolumn{2}{c}{  \textbf{GPT-4 \tutoreval}}  \\
& Presentation & Correctness & Presentation & Correctness \\
\midrule
Vicuna-13B-V1.5                  & 51.0 & 49.0 & 44.5 & 32.9 \\
\textsc{Llemma-7B-32K-Ultrachat} & 51.4 & 59.4 & 46.2 & 39.4 \\
\modelsmall                      & 67.8 & 67.9 & 62.4 & 50.0 \\
GPT-4                            & 92.3 & 91.9 & 89.5 & 85.2 \\         
\bottomrule
\end{tabular}
\vspace{10pt}

\label{tab:human_eval}
\end{table*}

\subsection{LM as an Evaluator}\label{sub:llm_evaluator}
\paragraph{Key points as ground-truths}
We ask human annotators to  provide a ground-truth solution to each question by sketching the \emph{key points} that a good teacher should address in answering the student. These key points  make it possible to quickly assess the correctness of an answer. Annotators were instructed to draw on their own experience as educators when compiling these key points.

\paragraph{LM-powered evaluation} At evaluation time, the LM tutor is prompted to answer each question based on the associated chapter. We let the LM generate up to 800 tokens.
To rate the free-form model generations, we prompt an LM evaluator to assign grades to each output. The LM evaluator is instructed to follow the key points as a guide for judging the answer. We use the January 2024 version of GPT-4-Turbo \cite{gpt4}. GPT-4 is prompted to grade  the output's \emph{presentation} and \emph{correctness}. \emph{Presentation} measures clarity of explanation  and general form. \emph{Correctness} tracks factual mistakes and incorrect references to the  chapter. 
Correctness is a more meaningful metric to benchmark LMs, so we usually report only the correctness scores. Nevertheless, presentation scores are informative, and we show that prompting GPT-4 for a presentation score improves human-GPT-4 correlation on correctness.

\begin{figure}[ht]
    \centering
    \includegraphics[width=0.75\linewidth]{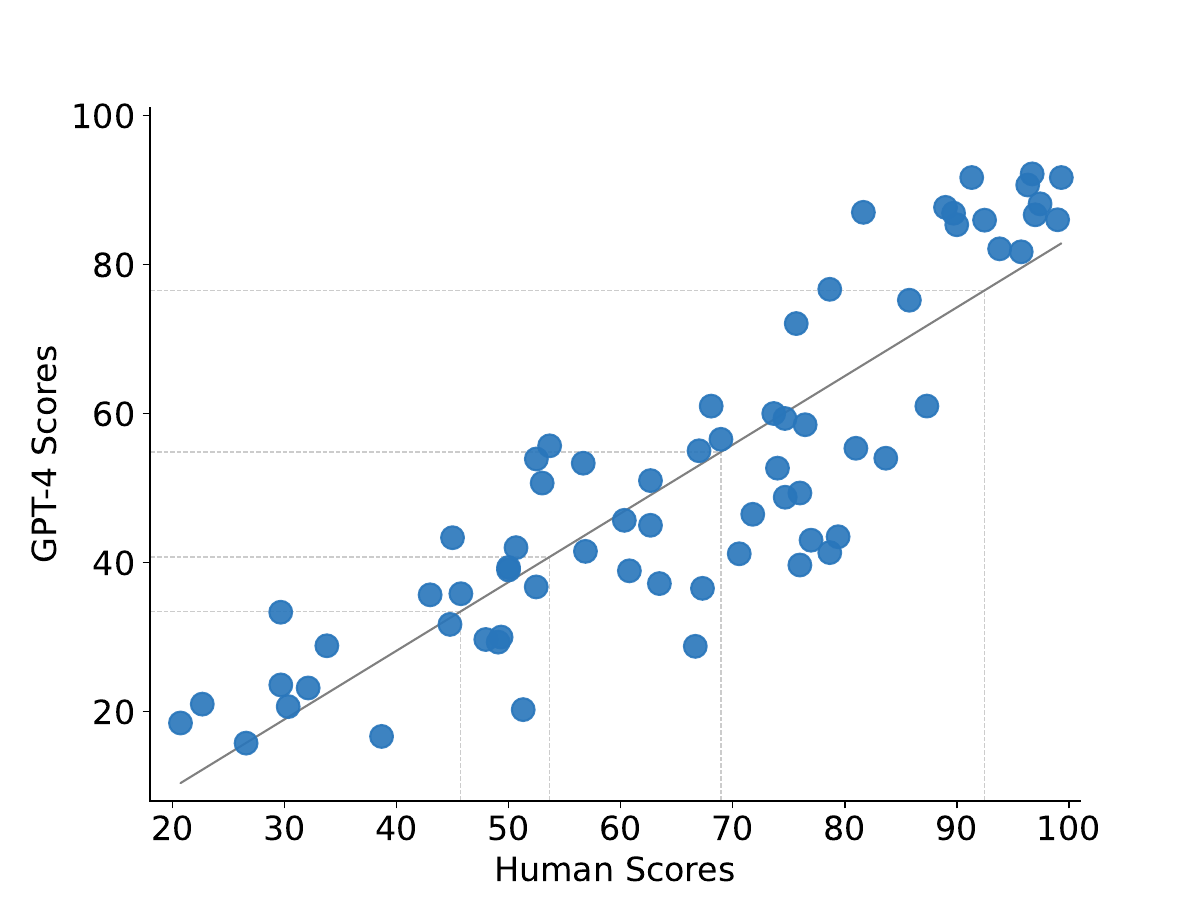}
    \caption{
        We show the correlation between the scores from 17 annotators   and the GPT-4 scores for four models: Vicuna-13B-16K, Llemma-7B-32K-Ultrachat, Llemma-7B-32K-MathMix, and GPT-4. Each annotator evaluates these models on their own set of 50 questions.
    }
    \label{fig:human_gpt4_corr}
\end{figure}

\paragraph{Human-GPT-4 agreement} To show that GPT-4 can be used as a reliable evaluator, we ask each human annotator to grade the outputs of four anonymized models on the 50 questions they created. We report the final scores from humans and GPT-4 in Table \ref{tab:human_eval}. 
We also compute 3 human-GPT-4 correlation metrics: the Pearson, the Spearman, and the Kendall-$\tau$ coefficients (more details in Appendix \ref{appendix:grading_analysis}). Results show that GPT-4 rankings correlate well with humans  rankings on  individual questions. 
Moreover, we find that  14/17 human annotators agree about the overall ranking and that GPT-4 reaches this ranking on 12/17 of the question sets. In Figure \ref{fig:human_gpt4_corr} we see that the overall scores attributed by each human to each model correlate with the scores obtained from GPT-4 on the same questions.

\paragraph{Additional analysis} We run ablations to further analyse the effect of our GPT-4-grading strategy on human-GPT-4 correlation. We consider the following modifications:
\begin{enumerate}
    \item Remove the key points from GPT-4's prompt
    \item Prompt GPT-4 for correctness without presentation
    \item Use GPT-3.5-Turbo instead of GPT-4-Turbo
\end{enumerate}

Correlation deteriorates in all cases, indicating the necessity of each component. See Appendix \ref{appendix:grading_analysis} for details, where we also show in Figure \ref{fig:human_gpt4_corr_gpt4only} that the key points are particularly useful for GPT-4 to grade its own generations fairly.

We also prompt GPT-4 for preference votes between various models. We find that GPT-4 preferences agree with the rankings obtained from individual grades. Hence our grading approach is at least as reliable as preference voting, while also being much less expensive. See Appendix \ref{appendix:grading_analysis}.

\paragraph{Future-proofing evaluation} Model updates to GPT-4 or using different evaluators will not preserve the \tutoreval scores.  
To facilitate future evaluations, we release the \tutoreval outputs of the models ranked in Figure \ref{fig:main_radar}, together with all GPT-4 and human grades.

\subsection{Evaluating State-of-the-art LMs on \tutoreval}\label{sub:leaderboard}

Figure~\ref{fig:main_radar} shows the leaderboard of \tutoreval with state-of-the-art LMs and their domain breakdown.
Detailed numbers can be found in Table \ref{tab:analysis_by_fields} in the appendix. We see in the radar plot that GPT-4 dominates in all domains followed by GPT-3.5 and Mixtral. 
Among 7B models, Mistral-7B-V2 is the strongest, closely followed by our \modelsmall, which we introduce in Section \ref{section:models}. Table \ref{tab:analysis_by_fields} shows that our model is stronger on the subset of math \tutoreval questions, owing to Llemma's specialized training on math data. 
 In Table \ref{tab:llama_comparisons}, we compare short-context models on \tutoreval-\textsc{ClosedBook}. 

As shown in Table \ref{tab:human_eval},  human evaluators give  92\% correctness to GPT-4 on \tutoreval. This confirms the fact that GPT-4 is already useful in educational settings. Future work will investigate how to build a harder academic benchmark. In the remainder of this paper, we show that \tutoreval performance for 7B and 34B models can be greatly improved with dedicated scientific training and fine-tuning.
\section{\dataset}\label{section:dataset}

In this section, we build the first dialogue dataset for science. We introduce \dataset, a dataset of high quality, long-form, synthetic conversations about textbook materials. In Section \ref{section:models}, we will show that \dataset is highly effective for improving performance on \tutoreval, compared to general-purpose dialogue datasets like UltraChat.

\subsection{Constructing \dataset}\label{sub:dataset_construction}

We construct \dataset in two steps: first we collect open-source textbooks which we process into a high-quality dataset of long documents, and secondly we generate dialogues about each textbook chapter by simulating teacher-student interactions with GPT-3.5-Turbo and GPT-4-Turbo.

\paragraph{Textbook collection} We scrape the domain libretexts.org for textbooks in all academic domains. All   textbooks are open-source. We note that libretexts.org is commonly found in web-scraped datasets, such as OpenWebMath. Our dataset, however, is exhaustive, structured, and cleaned. 

We collect the HTML from each textbook page and convert it to latex using the OpenWebMath HTML-to-LaTeX processor. We apply additional filtering to each page to remove bibliographies and metadata. To structure our dataset further, we concatenate short consecutive textbook chapters, so that each `chapter' in our dataset contains at least 512 words.
The resulting dataset consists in 1,685 textbooks, 78,000 chapters, and 160M words, and covers all education domains, including STEM, humanities and social sciences. The dataset is highly structured and documented, each chapter being traceable to its exact position in each textbook. We release this dataset for future research.

\paragraph{Dialogue generation} We generate \dataset  from our textbooks using GPT-3.5-Turbo \cite{gpt3} and GPT-4-Turbo \cite{gpt4}. We use two main prompting strategies: we simulate teacher-student conversations using two instances of GPT-3.5 and we use a single instance of GPT-4 to simulate an entire conversation.

For both generation strategies, we simulate \textit{open-book} and \textit{closed-book} conversations using textbook chapters. Open-book dialogues simulate  the student asking questions about the chapter to the teacher, and closed-book dialogues simulate the teacher using the chapter to give a lesson to the student about the chapter's topic. For closed-book dialogues, the teacher is prompted with the entire chapter but not the student, so that the resulting conversation can be understood without access to the textbook chapter.

Using GPT-4, we also generate \textit{textbook exams}: these are exam scripts, with solutions, based on a chapter,  without conversational formatting. We find that this helps GPT-4 include harder reasoning tasks than would normally arise in a conversational setting. See Appendix \ref{appendix:data_generation} for more details.

\dataset consists of 78K dialogues. On average, each dialogue contains 1,800 words over 15 turns. 40\% of the dataset is generated with GPT-4, and 60\% with GPT-3.5. 
We create a  validation split of 2.5K samples from \dataset. These textbook chapters were used by \tutoreval annotators to write questions.

\subsection{Dataset Inspection}
 \begin{figure}[t]
    \centering
    \includegraphics[width=0.75\columnwidth]{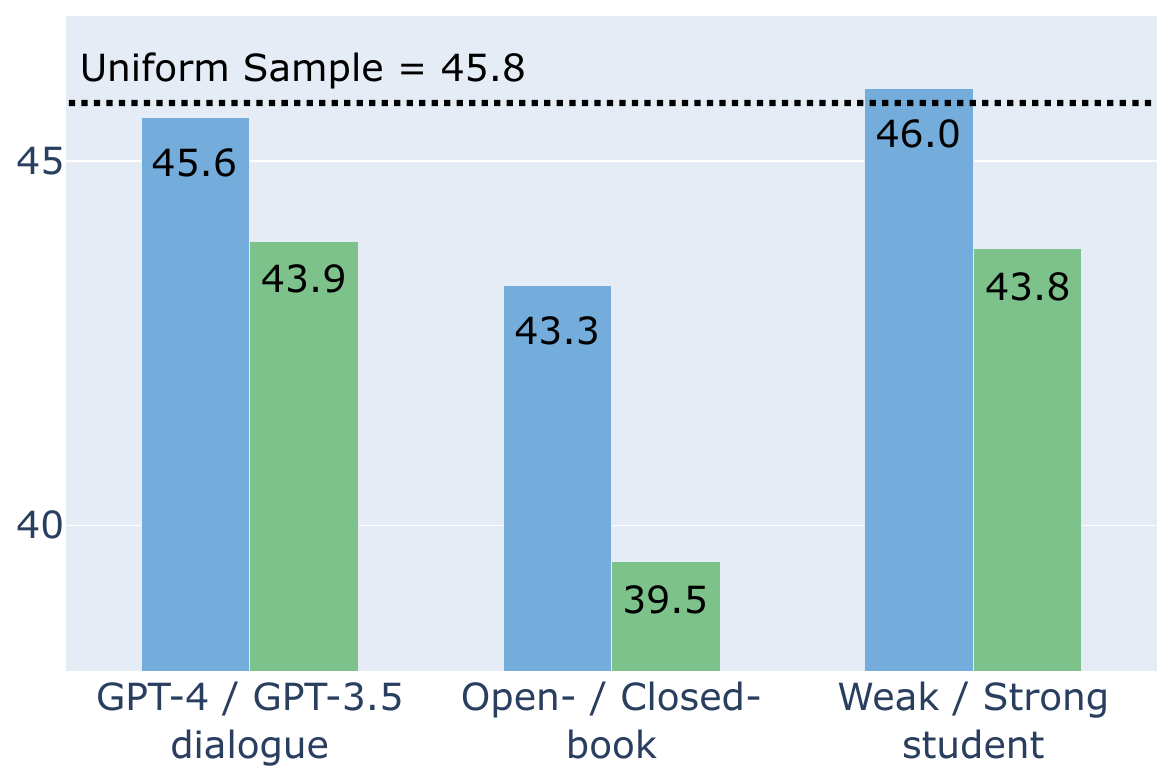}
    \caption{
        \tutoreval results for fine-tuning \textsc{Llemma-7B-32K} with various subsets of \dataset-STEM. Each subset contains 10K samples. See Table \ref{tab:data_ablations} for more  results.
    }
    \label{fig:compare_splits}
\end{figure}

\paragraph{Quality control} Empirically, we find that dialogues generated by GPT-4 are more factual and challenging than those written by GPT-3.5.  GPT-3.5 dialogues are more error-prone and sometimes wander, but they are more conversational, realistic, and friendly. See Appendix \ref{appendix:dialogue_examples} for examples of dialogues generated by GPT-3.5 and GPT-4.

\paragraph{\tutoreval contamination control} The train split of \dataset does not contain any chapters used in \tutoreval. We further verify that fine-tuning with \dataset does not introduce test-set leakage by fine-tuning \textsc{Llemma-7B-32K} (see Section \ref{sub:finetuning_llemma}) on \dataset-STEM where we remove the entire textbooks that were used to construct \tutoreval. We find that this ablation performs similarly to the model fine-tuned on the full \dataset-STEM train set (see Table \ref{tab:llemma_ft_comparisons}). On \tutoreval, the ablation achieves 50.4 compared to 47.8. On \tutoreval-\textsc{ClosedBook} the ablation achieves 47.8 compared to 48.2.

\paragraph{Dataset experiments} We run ablations to understand the impact of the \dataset generation strategy. We sample 10K sequences of the following types:
\begin{enumerate}
    \item Dialogues generated by GPT-4 versus  GPT-3.5
    \item Open-book dialogues versus closed-book
    \item Dialogues simulating a strong versus a weak student. Strong students are prompted to be confident and knowledgeable, whereas weak students are prompted to be unfocused, error-prone, and 
    \item A random sample of \dataset dialogues
\end{enumerate}
We fine-tune \textsc{Llemma-7B-32K} (see Section \ref{sub:finetuning_llemma}) on each dataset and we report results in Figure \ref{fig:compare_splits} and  Table \ref{tab:data_ablations} in Appendix \ref{appendix:data_ablations}. See also Appendix \ref{appendix:training_details} for data-processing details. On \tutoreval, we find that the GPT-4 data gives better results than  GPT-3.5 data. However, this performance gap is mostly due to the 13\% of `misleading' questions. Similarly, dialogues simulating weak students mostly help with misleading questions.  This shows that \dataset helps mitigate sycophancy \cite{sycophancy}. 

Open-book dialogues are also more effective than closed-book dialogues, both for \tutoreval and for \textsc{ClosedBook}. We believe that dialogues where an LM and a user discuss a common piece of text occur rarely in existing  datasets, making \dataset a valuable contribution.

\begin{table}[ht]
\centering
\caption{\tutoreval results comparing different pre-trained models and base models. Most pre-trained models are not able to process long contexts so we only evaluate them on \tutoreval-\textsc{ClosedBook}.}
\begin{tabular}{ lrr}
\toprule                   &   \textsc{Tutor}-           &  \textsc{Closed}-   \\ 
                          &    \textsc{Eval}  & \textsc{Book}\\    \midrule
\multicolumn{3}{l}{\textbf{Math-specialized Models}} \\
\midrule
MAmmoTH-7B       &    -           &  1.5        \\
MAmmoTH-13B      &   -            & 24.2        \\
MAmmoTH-70B      &    -           &  2.8       \\
WizardMath-7B    &  -             & 20.1        \\
WizardMath-13B   &  -             & 28.2        \\
WizardMath-70B   &  -             & 0.5         \\
Llama-Pro-8B     & -              & 31.6       \\
\midrule
\multicolumn{3}{l}{\textbf{Generalist Models}} \\
\midrule
Phi-2            &   -            & 28.6      \\
Tulu-2-DPO-7B    &  -             & 33.4        \\
Tulu-2-DPO-13B   &  -             & 42.6        \\
Tulu-2-DPO-70B   &  -             & 59.5     \\
Vicuna-13B-16K   & 33.2           & 37.7      \\
Mistral-7B-V1    & 30.9           & 36.8      \\
Zephyr-7B        & 46.3           & 50.4      \\
Mistral-7B-V2    & 50.9           & 53.8      \\
Mixtral-8x7B     & 66.3           & 68.2      \\
Llama-3-8B       & 55.3           & 59.1      \\
Llama-3-70B      & 71.3           & 78.3      \\
GPT-3.5-Turbo-1106    & 68.3      & 69.6      \\
GPT-4-Turbo-1106      & 85.2      & 86.1      \\
\midrule
\multicolumn{3}{l}{\textbf{Ablations}} \\
\midrule 
Llama-2-7B-32k-\mathdata &  30.8  & 31.7\\
CodeLlama-32k-\mathdata  &  43.2   & 37.1 \\
Llemma-7B-\mathdata      &  46.4  & 46.4       \\   
\midrule
\multicolumn{3}{l}{\textbf{Ours}} \\
\midrule 
\modelsmall               &   50.5       &  46.6      \\
\modellarge               &   57.2       &  57.6      \\ 
\bottomrule
\end{tabular}

\label{tab:llama_comparisons}
\end{table}


Finally, Table \ref{tab:data_ablations} shows that the random data sample has the most well-rounded performance across open-book and closed-book evaluations, and misleading and non-misleading questions. This shows that data diversity is key and that the engineering effort put into constructing \dataset has produced a useful dataset.

\section{Building LM Tutors}\label{section:models}

In this section, we investigate the main ingredients required for building strong and helpful LM tutors. In the following, we focus on developing LM tutors for math. Therefore, we ask the two following questions:

(1) How important are training with scientific texts and \dataset as drivers of performance on \tutoreval?

(2) How can we train an LM tutor that also has strong math problem-solving skills? 

In Section \ref{sub:finetuning}, we begin by introducing our two best models: \modelsmall and \modellarge, and their training recipes.
In Section \ref{sub:base_models}, \ref{sub:finetuning_llemma}, \ref{sub:math_tasks}, we answer the two above questions step-by-step, by  finding the most suitable base model and the best curated fine-tuning data. As a result, our best models achieve  competitive results on \tutoreval and math problem-solving tasks.

\subsection{Method}\label{sub:finetuning}

We present the two-stage fine-tuning pipeline which we use to obtain the best \tutoreval results. We first extend the LM's context-window to 32K tokens, and then train with \dataset.  See Appendix \ref{appendix:training_details} for hyperparameters.

\paragraph{Long-context training} Llemma-7B has a context window of 4K tokens. This is too short for \tutoreval, so we create a 5B-token dataset from OpenWebMath by sampling documents of up to 32K tokens. We fine-tune with an increased RoPE base \cite{rope}, following \citet{codellama}. We name this model \textsc{Llemma-7B-32K}. 
Because of computational limitations, we are not able to run this context-extension phase on Llemma-34B, but we still increase the RoPE base  before fine-tuning on \dataset.

\paragraph{\dataset fine-tuning} We process  teacher/student dialogues as  assistant/user dialogues. We fine-tune on open-book dialogues by including the entire textbook chapter, and on closed-book dialogues by removing the textbook chapter. See Appendix \ref{appendix:training_details} for processing details.

\paragraph{\mathdata: a dataset for LM math tutors} In Table  \ref{tab:llama_comparisons}, we compare fine-tuning  with  \dataset and  \dataset-STEM, the subset of dialogues about science textbooks. 
We find that we can further improve \tutoreval performance by combining \dataset-STEM and MetaMath. We process the original MetaMath dataset by randomly concatenating 10 question/answer pairs. The resulting MetaMath dataset contains 40K samples and an average sequence length of 3,000 tokens, similar to \dataset-STEM. We refer to the mix of \dataset-STEM and MetaMath as \mathdata.
We collect \tutoreval results in Tables \ref{tab:llama_comparisons} and \ref{tab:llemma_ft_comparisons}. See Sections \ref{sub:base_models}, \ref{sub:finetuning_llemma}, and \ref{sub:math_tasks} for analysis.

\begin{table}[ht]
\caption{\tutoreval  results for fine-tuning  \textsc{Llemma-7B-32K} with different dialogue datasets. We add MetaMath to each dataset because we are also interested in achieving strong performance on MATH and GSM8K, see Figure \ref{fig:tutoreval_gsm8k_math}. }
\centering
\begin{tabular}{ lcc}
\toprule                   &     \textsc{Tutor-}     & \textsc{Closed}-        \\       
Fine-tuning data                         & \textsc{Eval} & \textsc{Book}   \\    
\midrule
MetaMath                   &   14.5   &  20.5         \\

\midrule
UltraChat                 &   39.9   &  41.2    \\
~~~~+ MetaMath       &   35.0   &  45.1     \\
\midrule
\dataset~~~~~~~                   &   48.4  &  \textbf{50.9}           \\
~~~~+ MetaMath        &   48.2  &  49.7   \\
\midrule
\dataset-STEM     &   47.8   &  48.2               \\
 ~~~~+ MetaMath   ($=$\mathdata)              &   \textbf{50.5}    &  46.6     \\
\bottomrule
\end{tabular}

\label{tab:llemma_ft_comparisons}
\end{table}

\subsection{Investigating Foundation Models}\label{sub:base_models}
By comparing Llama-2 based models, CodeLlama and Llemma-7B, we find that \tutoreval benefits significantly from training on scientific text. 

\paragraph{Baselines} We fine-tune Llama-2-7B and CodeLlama-7B with \mathdata as in   Section \ref{sub:finetuning}. Recall that Llemma was trained from  CodeLlama, which was trained from Llama-2, so this helps isolating the effect of continued-training. To show the impact of long-context training, we also fine-tune a base Llemma model with \textsc{\mathdata} directly, and refer to this model as \textsc{Llemma-7B-\mathdata}.
Finally, we compare these models with MAmmoTH and WizardMath up to 70B parameters, Llama-Pro-8B-Instruct \cite{llamapro}, and various pre-trained models. The short-context models are only evaluated on \tutoreval-\textsc{ClosedBook}. 

\paragraph{Results} See Table \ref{tab:llama_comparisons}. Llama-2-based models do poorly on \tutoreval-\textsc{ClosedBook}. Only Tulu-2-DPO-70B improves over our best 7B model.
The poor scores achieved by the math models  confirms that training directly for GSM8K and MATH does not produce usable models. Note also that Llama-Pro-8B-Instruct was trained on the same data as Llemma and was tuned on a dataset including  MetaMath. Comparing with \textsc{Llemma-7B-MathMix}  shows the strength of our data.

In our experiments,  \textsc{CodeLlama-32K-\mathdata} ranks between our Llama-2  and Llemma models. This confirms the usefulness of training on code and on scientific data. Unsurprisingly, Table \ref{tab:analysis_by_fields} shows that CodeLlama  is strong on computer science questions.
Long-context training is found to improve \tutoreval performance by 4 points without harming \textsc{ClosedBook}, verifying that this helps long context understanding. 
In summary, comparing \modelsmall with other Llama-2 models shows the strength of training with scientific data and the efficiency of our fine-tuning. 
Table \ref{tab:analysis_by_fields} shows that \modelsmall is  stronger than Mistral-7B-V2 on math questions but weaker in other domains. Overall, Mistral performs slightly better than our best 7B Llemma model\footnote{Note that  Mistral-7B-V2 is an industry-grade closed-source LM and that \modelsmall is obtained from Llama-2 with open-source data.}. 

\subsection{Investigating Fine-tuning Datasets}\label{sub:finetuning_llemma}

Having established that \textsc{Llemma-7B-32K} is a strong base model, we compare different fine-tuning datasets.

\paragraph{Baselines} We fine-tune  \textsc{Llemma-7B-32K} with the following datasets:
\begin{enumerate}
    \item UltraChat (80K samples)
    \item \dataset (75K samples)
    \item \dataset-STEM (40K samples)
\end{enumerate}
We also investigate the effect of mixing MetaMath into each of these datasets. Recall that \mathdata is \dataset-STEM + MetaMath and consists of 80K samples total. 
For exact comparison with \mathdata, we only mix 40K UltraChat samples with MetaMath, giving 80K samples total.

\paragraph{Results} Table \ref{tab:llemma_ft_comparisons} shows that fine-tuning with \dataset  improves \tutoreval  by 9 points over UltraChat, clearly showing the benefits of \dataset.
\mathdata is the  strongest model on \tutoreval but is weaker than the \dataset models for closed-book evaluation. However, in Section \ref{sub:math_tasks}, we show that \mathdata is much stronger on math problem solving tasks.

\subsection{Investigating LM Tutors for Math}\label{sub:math_tasks}
\begin{figure}[t]
    \centering
    \includegraphics[width=0.99\columnwidth]{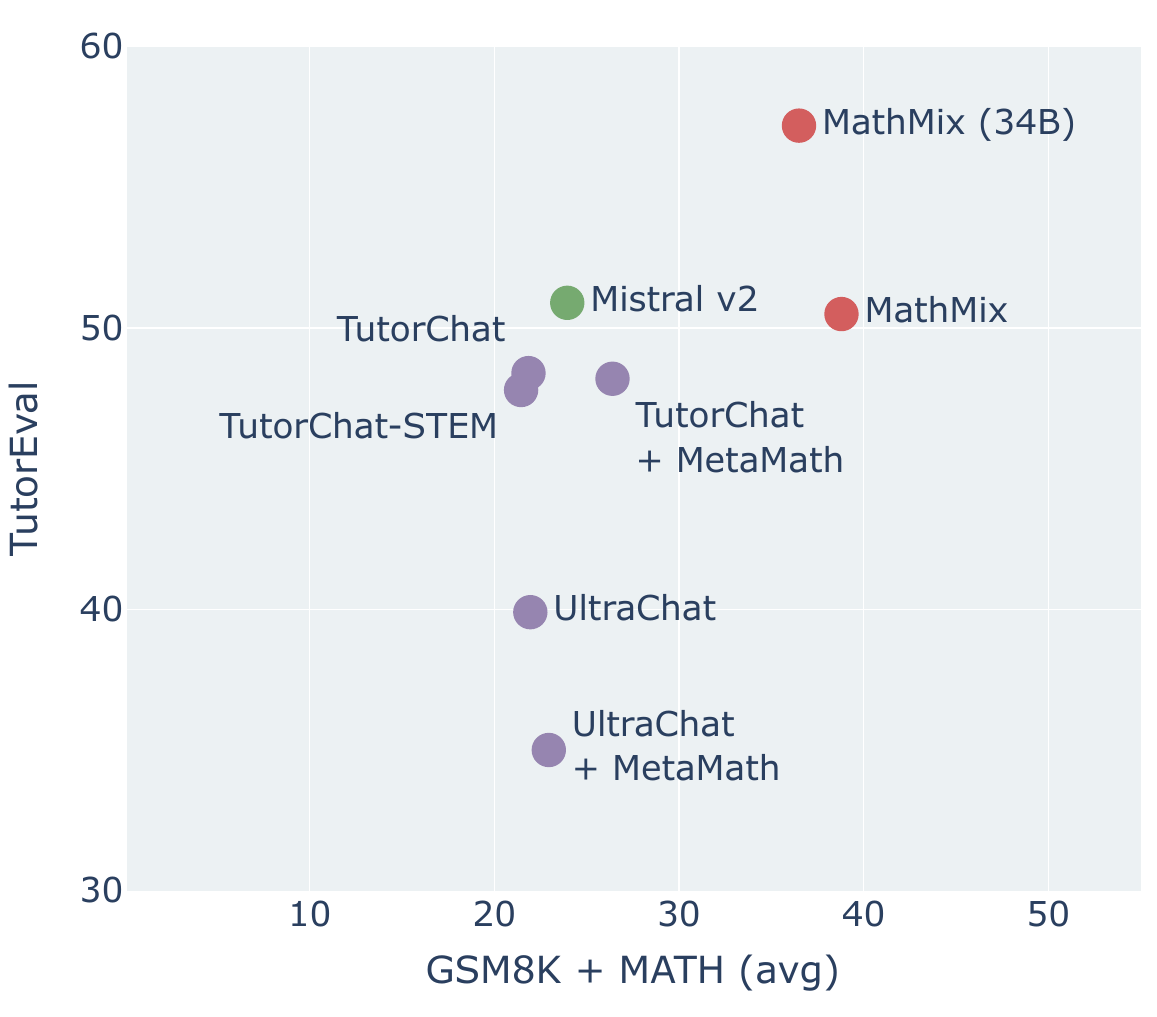}
    \caption{
        Combined performance on \tutoreval and math oriented datasets (average of GSM8K \& MATH). In red are our models trained with \mathdata, with 7B and 34B parameters. In purple are 7B-parameter baselines trained from Llemma-7B-32K. We also include the pre-trained  Mistral-7B-V2 in green.
    }
    \label{fig:tutoreval_gsm8k_math}
\end{figure}

Having identified that training with scientific texts and fine-tuning with \dataset improve \tutoreval performance significantly, we investigate how to build a strong LM tutor specialized in math.

In Figure \ref{fig:tutoreval_gsm8k_math}, we plot the \tutoreval performance of different fine-tuned Llemma-7B-32K models against their average  GSM8K and MATH scores (details in Table~\ref{tab:math_evals}).
We see that fine-tuning with \dataset alone does not improve performance on math tasks. Surprisingly, combining \dataset with MetaMath does not significantly improve math performance either. %
However, \dataset allows us to target science-specific dialogues, and we find that \dataset-STEM significantly improves math performance when combined with MetaMath to form \mathdata. 

The resulting model, \modelsmall, outperforms the strong Mistral-7B-V2 baseline on math tasks, while matching its \tutoreval performance. This strong result illustrates the richness and usefulness of the \dataset dataset. 
In Tables \ref{tab:code_evals} and \ref{tab:general_evals}, we also evaluate our models on code and a suite of general NLP tasks. We find that \modelsmall achieves similar coding performance to CodeLlama-7B.

\subsection{Human Evaluation and Empirical Observations}\label{sub:humaneval}

See Table \ref{tab:human_eval} for the human \tutoreval scores attributed to \textsc{Llemma-7B-32K-UltraChat} and \modelsmall. There is high agreement  about these model's rankings: 16/17 annotators rank  \textsc{MathMix} above  UltraChat  for correctness and presentation. 

Empirically, we find that \modelsmall tends to give very detailed answers. It often copies GPT-4's style by using enumerations and choosing similar examples. It also asks lots of follow-up questions.

\tutoreval does not evaluate multi-turn conversations. However, we observed that \modelsmall has the ability to hold a conversation about a given chapter over multiple turns without losing coherence. We also noticed that it adjusts the level of its answers depending on how confused the student appears to be, and it often picks up on the student answers by emphasizing how well the student understood the chapter.

We also observed that \modelsmall is quite robust to misleading questions, such as questions implying that the textbook is wrong. In such cases,  \modelsmall does not side with the student and politely but firmly clarifies the confusion.

\section{Conclusion and Future Work}

We have introduced \tutoreval and \dataset with the hope of encouraging the development of LMs as useful scientific assistants. Moreover, the methodology behind \tutoreval opens up new  avenues for creating ever-more specialised evaluations, relying on LM evaluators to measure behaviors not captured by other NLP metrics.

As follow-up work, we plan to keep building up \tutoreval to cover more science disciplines, more levels of difficulty, and more diverse use-cases. We believe that the dataset presented in this paper can be the foundation of a large-scale crowd-sourcing effort for the research community to define how it would like LMs to be applied as a transformative technology.

\section*{Impact Statement}
This paper presents work whose goal is to advance the field of Machine Learning. There are many potential societal consequences of our work, none of which we feel must be specifically highlighted here.

\section*{Acknowledgments}
ARF was supported by the Israel Academy of Sciences and Humanities \& Council for Higher Education Excellence Fellowship Program for International Postdoctoral Researchers and partially supported by STRANO PID2021-122752NB-I0.
SM acknowledges funding provided by the Sivian Fund and the Roger Dashen Member Fund at the Institute for Advanced Study.
This material is based upon work supported by the U.S. Department of Energy, Office of Science, Office of High Energy Physics under Award Number DE-SC0009988.

\bibliography{bibliography}
\bibliographystyle{icml2024}

\newpage
\appendix
\onecolumn

\section{Examples of \tutoreval questions}\label{appdenix:eval_examples}
Below we list some examples of \tutoreval questions by domain. 

\subsection{Mathematics} 

\textbf{Chapter url}:\\
\href{https://math.libretexts.org/Bookshelves/Algebra/Advanced_Algebra/09%3A_Sequences_Series_and_the_Binomial_Theorem/9.04%3A_Binomial_Theorem}{https://math.libretexts.org/Bookshelves/Algebra/Advanced\_Algebra/09\%3A\_Sequences\_Series\_and\_the\_Binomial\_Theorem/\\9.04\%3A\_Binomial\_Theorem} \\
\textbf{Question:} Can you walk me through the calculations for Exercise 1? I'd like to see another example worked out in the same level of detail as Example 2.

\textbf{Key points:}
\begin{itemize}
\item the question asks to calculate 8 choose 5
\item the answer is 56, as stated in the chapter
\item show the student how to use cancellations in the binomial coefficient formula to avoid carrying out large multiplications
\end{itemize}

\textbf{Additional information}
\begin{itemize}
    \item closed-book: no
    \item answer in chapter: yes
    \item misleading: no
    \item difficulty: easy
\end{itemize}

\subsection{Computer science}
\textbf{Chapter url:}
\href{https://eng.libretexts.org/Bookshelves/Computer_Science/Applied_Programming/Book%3A_Neural_Networks_and_Deep_Learning_(Nielsen)/06%3A_Deep_Learning/6.04%3A_Recent_progress_in_image_recognition}{https://eng.libretexts.org/Bookshelves/Computer\_Science/Applied\_Programming/Book\%3A\_Neural\_Networks\_\\and\_Deep\_Learning\_(Nielsen)/06\%3A\_Deep\_Learning/6.04\%3A\_Recent\_progress\_in\_image\_recognition}

\textbf{Question:} Is the dataset from ILSVRC-2012 a subset of the ImageNet dataset?

\textbf{Key points:}
\begin{itemize}
    \item the full name of ILSVRC-2012 is ImageNet Large Scale Visual Recognition Challenge 2012
    \item ILSVRC-2012 contains 1000 categories and 1.2 million images
\end{itemize}

\textbf{Additional information}:
\begin{itemize}
    \item closed-book: yes
    \item answer in chapter: yes
    \item misleading: no
    \item difficulty: easy
\end{itemize}

\subsection{Physics}

\textbf{Chapter url:}

\href{https://phys.libretexts.org/Bookshelves/Classical_Mechanics/Classical_Mechanics_(Dourmashkin)/13%3A_Energy_Kinetic_Energy_and_Work/13.05%3A_Work_done_by_Non-Constant_Forces}{https://phys.libretexts.org/Bookshelves/Classical\_Mechanics/Classical\_Mechanics\_(Dourmashkin)/13\%3A\_Energy\_\\Kinetic\_Energy\_and\_Work/13.05\%3A\_Work\_done\_by\_Non-Constant\_Forces}

\textbf{Question:} So the work done by a force F between x and y is simply the integral of F along the segment [x;y]?

\textbf{Key points:}

\begin{itemize}
    \item The statement is imprecise.
    \item It is the integral of the component of F in the $y-x$ direction.
\end{itemize}

\textbf{Additional information:}

\begin{itemize}
    \item closed-book: yes
    \item answer in chapter: yes
    \item misleading: yes
    \item difficulty: easy
\end{itemize}

\subsection{Life sciences}

\textbf{Chapter url:}
\href{https://med.libretexts.org/Bookshelves/Anatomy_and_Physiology/Anatomy_and_Physiology_1e_(OpenStax)/Unit_3%3A_Regulation_Integration_and_Control/17%3A_The_Endocrine_System/17.03%3A_The_Pituitary_Gland_and_Hypothalamus}{https://med.libretexts.org/Bookshelves/Anatomy\_and\_Physiology/Anatomy\_and\_Physiology\_1e\_(OpenStax)/\\Unit\_3\%3A\_Regulation\_Integration\_and\_Control/17\%3A\_The\_Endocrine\_System/17.03\%3A\_The\_Pituitary\_Gland\_and\\\_Hypothalamus}

\textbf{Question:} The section on antidiuretic hormone states that dehydration can result in high blood osmolarity. I don't really understand why that would happen. Will you please explain it to me?

\textbf{Key points:}

\textbf{Key points:} 
\begin{itemize}
\item The student is confused about the relationship between blood osmolarity and hydration
\item State that osmolarity is proportional to the amount of dissolved particles per unit volume
\item Mention that the volume of consumed water influences blood volume
\item State that lower blood volume given the same number of dissolved particles must result in higher osmolarity
\end{itemize}

\textbf{Additional information:} 
\begin{itemize}
    \item closed-book: no
    \item answer in chapter: no 
    \item misleading: no
    \item difficulty: easy
\end{itemize}

\subsection{Environmental science}

\textbf{Chapter url:}

\href{https://eng.libretexts.org/Bookshelves/Environmental_Engineering_(Sustainability_and_Conservation)/Book%3A_Essentials_of_Environmental_Science_(CK-12)/13%3A_Untitled_Chapter_13/13.03%3A_New_Page}{https://eng.libretexts.org/Bookshelves/Environmental\_Engineering\_(Sustainability\_and\_Conservation)/Book\%3A\_Essentials\\\_of\_Environmental\_Science\_(CK-12)/13\%3A\_Untitled\_Chapter\_13/13.03\%3A\_New\_Page}

\textbf{Question:} So eutrophication could lead to hypoxia due to explosive growth of algae that consume oxygen?

\textbf{Key points:}
\begin{itemize}
    \item This is false.
    \item The student is confused about the causes of oxygen depletion.
    \item Need to clarify the logical chain of eutrophication.
\end{itemize}

\textbf{Additional information}: 

\begin{itemize}
    \item closed-book: no
    \item answer in chapter: yes
    \item misleading: no
    \item difficulty: hard
\end{itemize}

\newpage
\section{Detailed \tutoreval results}\label{appendix:detail_results}

\begin{table*}[ht]
\caption{
    Detailed \tutoreval results for long-context models evaluated in this paper. We show scores by domain, difficulty, on the set of misleading questions, and in closed-book evaluation. Models are grouped as follows (top-down): (1) Strong pre-trained LLMs (2) Medium-sized pre-trained LMs (3) Our best models (4) Ablations of Llemma-7B-32K fine-tuned with different \dataset-based datasets (4) Ablations of  different base models fine-tuned with MathMix (5) Ablations of Llemma models fine-tuned without \dataset. 
}
\centering
\resizebox{\columnwidth}{!}{
\begin{tabular}{ l c c c c c | c c | c | c  c}
\toprule
& \multicolumn{5}{c|}{Domain} & \multicolumn{2}{c|}{Difficulty} & Mis- & \textsc{Tutor-} &  \textsc{Closed-} \\
Models & Math & Physics & CS & Life & Env  & Easy & Hard & leading &  \textsc{Eval} & \textsc{Book} \\ 
\midrule
GPT-4           & 83.4 & 84.0 & 88.9 & 91.3 & 86.5 & 88.6 & 77.7 & 75.9 & 85.5 & 87.5\\
Llama-3-70B     & 63.2 & 73.2 & 81.1 & 84.4 & 72.0 & 75.0 & 62.3 & 64.0 & 71.3 & 78.3 \\
GPT-3.5         & 63.4 & 65.9 & 77.1 & 79.6 & 75.8 & 73.1 & 58.3 & 56.6 & 68.8 & 70.3\\ 
Mixtral-8x7B    & 61.8 & 63.1 & 75.0 & 72.9 & 74.2 & 71.3 & 55.3 & 54.4 & 66.7 & 68.7\\ 
\midrule
Vicuna-13B-16K  & 25.6 & 28.5 & 47.2 & 46.3 & 36.8 & 38.1 & 21.3 & 23.4 & 33.2 & 37.7\\ 
Mistral-7B-V1   & 21.6 & 27.3 & 45.9 & 37.9 & 43.4 & 34.2 & 22.8 & 20.5 & 30.9 & 36.8\\ 
Zephyr-7B-Beta  & 36.7 & 43.9 & 60.5 & 53.8 & 59.1 & 50.5 & 35.9 & 33.5 & 46.3 & 50.4\\ 
Mistral-7B-V2   & 43.3 & 51.1 & 60.7 & 53.3 & 61.6 & 55.2 & 40.3 & 32.9 & 50.9 & 53.8\\ 
Llama-3-8B      & 46.5 & 52.9 & 68.2 & 68.1 & 61.0 & 59.8 & 44.1 & 43.6 & 55.3 & 59.8 \\
\midrule
Llemma-7B-32K-MathMix               & 45.7 & 51.1 & 56.8 & 55.0 & 53.1 & 56.6 & 35.6 & 29.4 & 50.5 & 46.6\\ 
Llemma-34B-MathMix                  & 52.2 & 58.8 & 65.2 & 55.4 & 56.9 & 62.0 & 45.4 & 40.2 & 57.2& 57.6\\ 
\midrule
Llemma-7B-32K-TutorChat-STEM        & 43.2 & 48.6 & 55.0 & 49.2 & 48.4 & 52.6 & 36.2 & 29.0 & 47.8 & 48.2\\
Llemma-7B-32K-TutorChat             & 45.3 & 48.6 & 55.0 & 46.3 & 45.0 & 54.0 & 34.7 & 27.8 & 48.4 & 50.9 \\
Llemma-7B-32K-TutorChat+MetaMath    & 43.7 & 47.0 & 56.0 & 49.2 & 51.6 & 53.5 & 35.1 & 28.5 & 48.2 & 49.7\\
\midrule
Llama-32K-MathMix                   & 21.0 & 33.9 & 40.0 & 53.8 & 34.3 & 33.5 & 24.2 & 23.5 & 30.8 & 31.7\\
CodeLlama-32K-MathMix               & 38.4 & 38.6 & 56.3 & 42.1 & 41.2 & 48.4 & 30.4 & 30.7 & 43.2 & 37.1\\ 
Llemma-7B-MathMix                   & 42.9 & 45.3 & 55.0 & 42.5 & 44.3 & 51.9 & 33.1 & 26.5 & 46.4 & 46.4\\ 
\midrule
Llemma-7B-32K-MetaMath              & 14.4 & 14.0 & 17.8 & ~~9.2 & ~~7.6 & 16.3 & ~~9.9 & ~~7.2 & 14.5 & 20.5 \\ 
Llemma-7B-32K-Ultrachat             & 34.3 & 37.4 & 50.8 & 41.3 & 43.1 & 44.6 & 28.2 & 22.1 & 39.9 & 41.2\\ 
Llemma-7B-32K-MetaMath-Ultrachat    & 29.8 & 29.9 & 47.9 & 32.1 & 39.6 & 38.1 & 25.4 & 21.2 & 35.0 & 45.1\\
\bottomrule
\end{tabular}
}
\label{tab:analysis_by_fields}
\end{table*}


\begin{table*}[ht]
\caption{
    Detailed human  \tutoreval results. The 4 models below are graded blind by all annotators who contributed to \tutoreval. Each annotator grades the model on their set of approximately 50 questions and we collect all results to compute the final \tutoreval scores. 
}
\centering
\begin{tabular}{ l c c c c c | c c | c | c }
\toprule
& \multicolumn{5}{c|}{Domain} & \multicolumn{2}{c|}{Difficulty} & Misleading & \textsc{Tutor-} \\
Models & Math & Physics & CS & Life & Env  & Easy & Hard & Questions &  \textsc{Eval}  \\ 
\midrule
GPT-4                               & 88.6 & 92.2 & 94.5 & 95.8 & 93.8 & 93.6 & 85.7 & 82.7 & 91.2 \\
Llemma-7B-32K-MathMix               & 59.6 & 74.5 & 77.4 & 62.1 & 72.5 & 71.6 & 52.3 & 48.1 & 66.5 \\ 
Llemma-7B-32K-Ultrachat             & 51.6 & 64.3 & 69.6 & 45.4 & 53.2 & 62.3 & 41.7 & 39.3 & 56.8 \\ 
Vicuna-13B-16K                      & 51.4 & 37.9 & 67.0 & 43.8 & 49.1 & 53.6 & 34.0 & 37.0 & 48.5 \\ 
\bottomrule
\end{tabular}
\label{tab:expanded_human_eval}
\end{table*}

\begin{table*}[ht]
\caption{
    We compare the  GPT-4 and Llemma-34B-MathMix scores on different subsets of \textsc{TutorEval} questions, evaluated both open-book (\tutoreval) and closed-book (\textsc{ClosedBook}). Note that the final \textsc{ClosedBook} score is the score achieved on closed-book questions only (370 questions) but that \tutoreval includes all questions (834 questions). These results show that open-book questions cannot be evaluated with \textsc{ClosedBook}. These results also show how well LMs are able to utilize the added context to extract the answer when the question is marked as \emph{answer in chapter}.}
\centering
\resizebox{0.9\columnwidth}{!}{
\begin{tabular}{lccccc}
\toprule
& \multicolumn{2}{c}{GPT-4}                       & \multicolumn{2}{c}{Llemma-34B-MathMix}     & Number of \\
& \textsc{TutorEval}        & \textsc{ClosedBook} & \textsc{TutorEval} & \textsc{ClosedBook}   &   Questions \\
\toprule
Open-book Questions & 83.4 & 67.2 & 54.9 & 42.9 & 464\\
~~~~~~~~~~ Answer \textit{not} in Chapter & 81.4 & 67.1 & 50.0 & 40.5 & 261\\
~~~~~~~~~~ Answer in Chapter             & 89.6 & 63.0 & 60.0 & 40.1 & 203\\
\midrule
Closed-book Question & 88.0 & 87.5 & 60.1 & 57.6 & 370\\
~~~~~~~~~~ Answer \textit{not} in Chapter & 85.1 & 84.2 & 56.6 & 51.9 & 265\\
~~~~~~~~~~ Answer in Chapter             & 93.9 & 89.8 & 68.8 & 63.4 & 105\\
\midrule
Final score & 85.5 & 87.5 & 57.2 & 57.6 & 834/370\\
\bottomrule
\end{tabular}
}
\label{tab:gpt4_ablation}
\end{table*}

\newpage

\section{\tutoreval grading}\label{appendix:grading_analysis}
\subsection{Grading template}

\begin{tcolorbox}[colback=white, colframe=black, arc=4mm, boxrule=0.5mm]
Below is the template used for prompting the LM-evaluator to assign grades to model outputs on \tutoreval.

\medskip 

\small{
\texttt{Your task is to evaluate the teaching abilities of a new AI system which is interacting with a student about a science topic. The student and AI system are working together on a textbook chapter, and I would like you to rate how well the AI system addressed the student's question. \\ \\
You should give scores from 0 to 3 for PRESENTATION and CORRECTNESS. Half points are allowed. Please refer to the following descriptions: \\ 
\\
PRESENTATION: the AI provides an engaging response which will make the student want to learn more. Examples of good presentation skills include: giving the response a clear and helpful structure, picking up on positive aspects of the student's contributions, using examples to clarify complicated ideas, explaining complicated arguments in detail, adding follow-up and broadening remarks, etc. \\ 
CORRECTNESS: the AI correctly understands the question and the answer is true and accurate. The answer does not contain any false or misleading statements. The AI does not include any irrelevant information  and does not omit any essential reasoning steps. The AI also correctly relates the question to the chapter's content. Pay particular attention to reasoning and calculation mistakes. 
\\
Here is the textbook chapter used for this interaction: \\
\\
"""\\
PASSAGE\\
"""\\
\\
Here is the student's question: \\
\\
"""\\
QUESTION\\
"""\\
\\
To help you in your evaluation, we've compiled some ground-truth key points which a good AI system should cover in its answer. You MUST check that the AI's answer agrees with these key points. These key points have been checked by experts and are 100\% correct. These key points are particularly useful for spotting CORRECTNESS errors.\\
\\
"""\\
Key points to cover:\\
KEY POINTS\\
"""\\
\\
Here is the AI's answer:\\
\\
"""\\
ANSWER\\
"""\\
\\
Please present your scores as follows:\\
PRESENTATION: [explanation]. GRADE: x/3\\
CORRECTNESS: [explanation]. GRADE: x/3\\
}}
\end{tcolorbox}

\subsection{Grading ablations}

We collect statistics to help us understand how well \tutoreval correlates with  human for judging the 4 models evaluated in Table \ref{tab:human_eval}. For each of the 834 \tutoreval questions, we compute 3 statistics: the Pearson, the Spearman, and the Kendall-$\tau$ coefficient between GPT-4 and human grades. The Pearson measures how close GPT-4's grades are to the human grades, and the Spearman and Kendall-$\tau$ measure how similar the rankings are. When computing the Spearman and Kendall-$\tau$, we count it as 0 if either GPT-4 assigns same grades to all models while human grades vary, or if human grades are identical while GPT-4's grades differ.

We average each of these coefficients over the 834 questions. Correlation coefficients for correctness are reported in Table \ref{tab:human_gpt_correctness_correlation} and coefficients for presentation are in Table \ref{tab:human_gpt_presentation_correlation}. We compare the final values with the statistics we obtain in three other evaluation configurations: 
\begin{enumerate}
    \item Evaluate without showing GPT-4 the ground-truth key points
    \item Evaluate without prompting GPT-4 for a presentation grade
    \item Evaluate with GPT-3.5 instead of GPT-4
\end{enumerate}

\begin{figure*}[t]
    \centering
    \includegraphics[width=0.4\columnwidth]{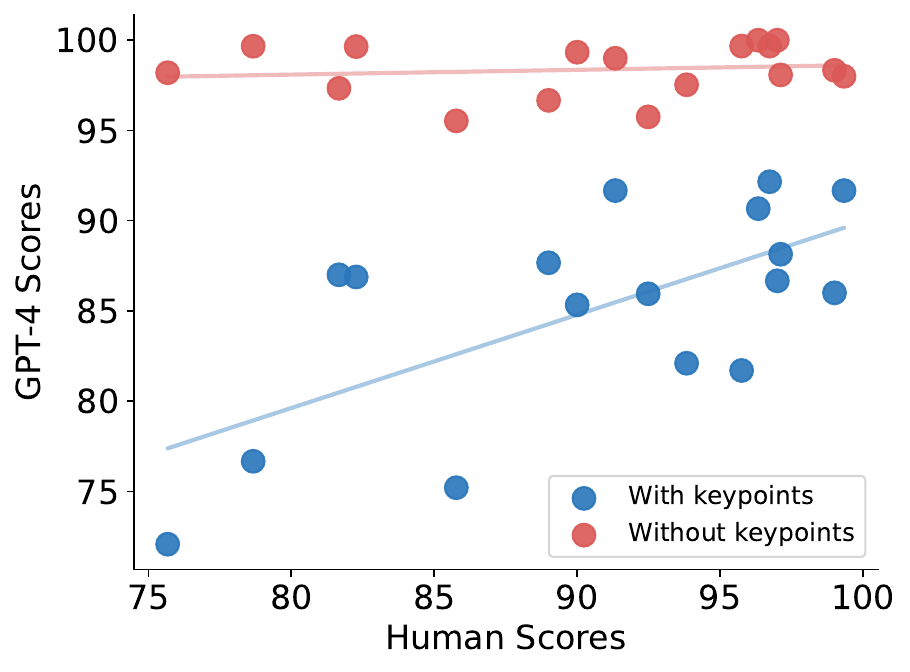}
    \caption{
        For each human annotator, we plot the score attributed to GPT-4 against the score attributed to GPT-4 by itself, with and without the key points. We see that the key points make it possible for  GPT-4 to grade its  own generations fairly.
    }
    \label{fig:human_gpt4_corr_gpt4only}
\end{figure*}

In all cases, we see that at least one correlation coefficient drops compared to the standard \tutoreval configuration. In Figure \ref{fig:human_gpt4_corr_gpt4only}, we see that the key points are particularly useful for GPT-4 to grade its own generations fairly.

\begin{table}[ht]
\centering
\resizebox{\linewidth}{!}{
\begin{tabular}{rcccc}
\toprule
Correctness \\

correlation &  \tutoreval grading & Grading without key-points & Grading without presentation & GPT-3.5 grading\\ 
 \midrule
Pearson & 60.0 & 56.7 &  56.4 & 47.3 \\
Spearman rank &58.1 & 55.1 & 58.0& 46.6  \\
Kendall-$\tau$ & 54.8  &  52.2 & 54.9 & 44.6 \\

\bottomrule
\end{tabular}
}
\caption{We report three correlation coefficients between the human and  LLM-evaluator correctness scores for each question in \tutoreval. The Pearson coefficient measures human/LLM correlation of grades, and the Spearman rank and Kendall-$\tau$ coefficients measure correlation of ranking. We run ablations on the standard \tutoreval approach by (1) grading with GPT-4 without the key points, (2) grading with GPT-4 without a presentation score and (3) grading with GPT-3.5.}
\label{tab:human_gpt_correctness_correlation}
\end{table}

\begin{table}[ht]
\caption{Three correlation coefficients for presentation scores between human and LLM-evaluator. }
\centering
\begin{tabular}{rccc}
\toprule
Presentation            \\
correlation     &  \tutoreval grading &    Grading without key points     &  GPT-3.5 grading \\ 
 \midrule
Pearson         & 69.2 &  68.2 & 37.1 \\ 
Spearman        & 67.6 & 67.2 & 36.5 \\
Kendall-$\tau$  & 63.2  & 62.8 & 34.2  \\
\bottomrule
\end{tabular}

\label{tab:human_gpt_presentation_correlation}
\end{table}

In order to compare our GPT-4-grading approach to \tutoreval with other evaluations using GPT-4 as an evalutor, such as \cite{alpaca_eval}, we compute GPT-4 preference votes for three pairs of models:
\begin{enumerate}
    \item \modellarge and Mistral-7B-V2
    \item Mistral-7B-V2 and \modelsmall
    \item \modelsmall and Llemma-7B-32K-UltraChat.
\end{enumerate}
For each pair of models A and B, we prompt GPT-4 with the outputs of the two models for each \tutoreval question twice, switching the order between the two models. We prompt GPT-4 to decide if Model A or B is more correct, or if there is a tie. When GPT-4 changes does not identify the same model during both rounds, we count this as a tie. 

Win  and tie rates are collected in Table \ref{tab:preferences}. We find that the ranking that results from preference votes coincides with the ranking of Table \ref{tab:llama_comparisons}, even for models which rank relatively closely, like Mistral-7B-V2 and \modelsmall. We conclude that grading is as reliable as preference voting, while being considerably cheaper and faster to run.

\begin{table*}[ht]
\centering
\begin{tabular}{ rc}
  \toprule                       & win / Tie / win (\%) \\
  \midrule
\modellarge                  &         \textbf{17.1}                  \\
           VS                   &       69.9                  \\
Mistral-7B-V2              &         12.9                  \\
\midrule
Mistral-7B-V2              &        \textbf{17.9 }         \\
            VS                  &         68.0                  \\
\modelsmall                  &         14.1                  \\
\midrule        
\modelsmall                  &         \textbf{21.4}                     \\
            VS                  &         70.0                  \\
Llemma-7B-32K-Ultrachat&      8.8                     \\
\bottomrule
\end{tabular}

\caption{Win-rate comparisons of various models on \tutoreval questions, judged by GPT-4. The preference-based ranking matches the gradings-based ranking  we obtain in Table \ref{tab:llama_comparisons}.}

\label{tab:preferences}
\end{table*}

\newpage
\section{\dataset composition}

\begin{table}[ht]
\caption{Summary statistics for \dataset dialogues}
\centering
\begin{tabular}{ l ccc }
\toprule
Domains & Textbooks & Dialogues & Words \\ 
\midrule
Biology         & 70  & 6K  & 9M \\ 
Business        & 77  & 7K  & 12M \\
Chemistry       & 119 & 12K & 20M \\
Engineering     & 134 & 6K  & 10M \\ 
Geology         & 39  & 2K  & 4M \\ 
Human sciences  & 556 & 13K & 21M \\ 
Mathematics     & 156 & 7K  & 12M \\ 
Medicine        & 91  & 4K  & 7M \\
Physics         & 71  & 5K  & 8M \\ 
Social sciences & 248 & 12K & 20M \\
Statistics      & 33  & 2K  & 3M \\ 
Workforce       & 91  & 2K  & 3M \\ 
\midrule
Total STEM            & 713 & 45K & 75M \\
Total           & 1685 & 78K & 130M \\ 
\bottomrule
\end{tabular}

\label{tab:dataset_composition}
\end{table}


\section{\dataset generation}\label{appendix:data_generation}

 We generate open-book conversations by simulating a student asking questions about this chapter. We  generate closed-book conversations by showing the chapter only to the teacher and prompting the teacher to conduct a class based on the chapter. We  diversify our templates by simulating different teacher and student personalities. For example, we sometimes prompt the student to be bright and keen or slow and badly behaved. In total we use 16 different templates.
 
 \dataset counts 15\% of textbook exams, and the rest of \dataset is evenly split between open-book and closed-book dialogues.

Below is an example of a template used for generating conversations for \dataset. We initialise the `student' and the `teacher' with system prompts, and we start off the conversation by using a fixed template. We include an entire textbook chapter into this template.

\begin{tcolorbox}[colback=white, colframe=black, arc=4mm, boxrule=0.5mm]
\textbf{Student system prompt} \\
\texttt{You are a student who struggles with staying focused and often finds yourself distracted during lessons. In your one-on-one tutoring session, you are dealing with a topic you're not very familiar with. You find yourself drifting off or not fully engaging with the material. Your lack of focus sometimes leads you to ask about things you've already covered or to miss key points in the explanation.}
\medskip

\textbf{Teacher system prompt} \\
\texttt{You are a strict and detail-oriented teacher. You insist on precision and thoroughness in all answers. Your teaching style is focused on ensuring students understand the minutiae of every topic. You engage with your students actively but always emphasize the importance of accuracy and detail in their responses. While you are knowledgeable and committed to student learning, your approach is rigid and exacting, leaving no room for ambiguity or imprecise answers.}

\medskip
\textbf{Template to start conversation} \\
\texttt{Here is a textbook chapter that you should learn for your upcoming exam. Read it carefully and let me know if you have any questions. This is your main opportunity to clarify any misunderstandings, so make the most of it! Don't be afraid to ask silly questions, I'm here to help!\\
\#\#\#\#\#\#\#\#\#\# \\
\{text\} \\ 
\#\#\#\#\#\#\#\#\#\# \\
Remember I'm your teacher so I'm here to answer all your questions about the textbook! Fire away!}
    
\end{tcolorbox}

\section{\dataset examples}\label{appendix:dialogue_examples}

Below we give two examples of dialogues in \dataset. First we give a dialogue generated by GPT-4 about a linear algebra textbook chapter, and secondly we give an example of a GPT-3.5-generated dialogue about a biology textbook chapter.

\subsection{GPT-4-generated dialogue}

The following dialogue was generated using GPT-4. GPT-4 was presented with the following textbook chapter: 
\href{https://math.libretexts.org/Bookshelves/Linear_Algebra/Interactive_Linear_Algebra_(Margalit_and_Rabinoff)/05%3A_Eigenvalues_and_Eigenvectors/5.03%3A_Diagonalization}{\textbf{{Click here}}} and was prompted to write a teacher-student dialogue simulating a class about its content.

<s>\\
assistant: Today we're going to discuss the concept of diagonalization of matrices. To start, let's consider a simple 2x2 matrix and determine whether it's diagonalizable. Take the matrix A = \(\left(\begin{array}{cc}3 & 1 \\ 0 & 2\end{array}\right)\). Can you find the eigenvalues of this matrix?</s> \\
user: Uh, I'm not sure how to find eigenvalues.</s> \\
assistant: No problem. To find the eigenvalues, we need to solve the characteristic equation \(det(A - \lambda I) = 0\), where \(I\) is the identity matrix and \(\lambda\) represents the eigenvalues. Can you calculate the determinant of \(A - \lambda I\)?</s> \\
user: I think so. I subtract \(\lambda\) from the diagonal entries, right? So it becomes \(\left(\begin{array}{cc}3-\lambda & 1 \\ 0 & 2-\lambda\end{array}\right)\).</s> \\
assistant: That's correct. Now find the determinant of that matrix.</s> \\
user: I multiply the diagonal entries and then subtract the product of the off-diagonal entries, which gives me \((3-\lambda)(2-\lambda) - (0)(1)\), so the determinant is \(\lambda^2 - 5\lambda + 6\).</s> \\
assistant: Almost there, but be careful. The determinant of a 2x2 matrix is the product of the diagonal entries minus the product of the off-diagonal entries. Since the off-diagonal entries in the lower left are zero, you only need to consider the diagonal product. So what's the determinant now?</s> \\
user: Oh, I see. So the determinant is just \((3-\lambda)(2-\lambda)\), which simplifies to \(\lambda^2 - 5\lambda + 6\).</s> \\
assistant: Exactly. Now solve for \(\lambda\) when the determinant equals zero.</s> \\
user: So I set \(\lambda^2 - 5\lambda + 6 = 0\) and solve for \(\lambda\). I think the solutions are \(\lambda = 2\) and \(\lambda = 3\).</s> \\
assistant: That's right, those are the eigenvalues of matrix A. Now, let's find the eigenvectors corresponding to each eigenvalue. Start with \(\lambda = 2\).</s> \\
user: I plug \(\lambda = 2\) back into the matrix \(A - \lambda I\), right? So I get the matrix \(\left(\begin{array}{cc}1 & 1 \\ 0 & 0\end{array}\right)\).</s> \\ 
assistant: Correct. Now find a vector \(v\) such that \((A - 2I)v = 0\).</s> \\ 
user: So I'm looking for a vector that makes this matrix equation equal zero... I'm not sure how to do that.</s> \\
assistant: Think of it as a system of linear equations. What vector can you multiply by that matrix to get the zero vector?</s> \\
user: Well, if I take \(v = \left(\begin{array}{c}-1 \\ 1\end{array}\right)\), then I get the zero vector when I multiply it by \(\left(\begin{array}{cc}1 & 1 \\ 0 & 0\end{array}\right)\).</s> \\
assistant: Exactly, so you found an eigenvector for \(\lambda = 2\). Now find an eigenvector for \(\lambda = 3\).</s> \\
user: I plug \(\lambda = 3\) into \(A - \lambda I\) and get \(\left(\begin{array}{cc}0 & 1 \\ 0 & -1\end{array}\right)\). So I need a vector that will give me the zero vector when I multiply... I'm not sure what to pick.</s> \\
assistant: Look at the system of equations you get from that matrix. It tells you that for any vector \(v\), the first component doesn't matter, but the second component must be the negative of the first to cancel out the \(1\). Can you find a vector that satisfies this condition?</s> \\
user: Oh, I think I get it. So if I choose \(v = \left(\begin{array}{c}1 \\ 0\end{array}\right)\), it satisfies the condition.</s> \\
assistant: Exactly right. Now that we have the eigenvalues and eigenvectors, we can check the criteria for diagonalizability. A matrix is diagonalizable if it has enough linearly independent eigenvectors. Since our matrix is 2x2, we need 2 linearly independent eigenvectors. Do we have that?</s> \\
user: Yes, the vectors \(\left(\begin{array}{c}-1 \\ 1\end{array}\right)\) and \(\left(\begin{array}{c}1 \\ 0\end{array}\right)\) are linearly independent.</s>

\subsection{GPT-3.5-generated dialogue}
<s>\\
assistant: Today, we will dive into the intricate world of chromatin, chromosomes, and DNA organization within the eukaryotic cell. We'll explore the distinctions between heterochromatin and euchromatin, the structure of chromosomes, the packing of DNA through supercoiling, the role of telomeres in DNA replication, and the fascinating mechanisms of chromatin loop formation. The goal is to gain a comprehensive understanding of these fundamental concepts in biochemistry.

Does this sound like a good plan for our session today? Can you follow along so far?</s>

user: I understand that you're covering some complex topics today, but I must admit that I'm not very familiar with them. I'll do my best to stay focused, but I may need some extra help to ensure I understand everything properly. Can we maybe start by revisiting the basics before delving into the more intricate details? That might help me grasp the concepts better.</s>

assistant: Absolutely, it's perfectly fine to revisit the basics before delving into the more complex details. Understanding the fundamentals is crucial for building a strong foundation in biochemistry. Let's start by revisiting the basics.

First, let's discuss the distinction between heterochromatin and euchromatin. Heterochromatin and euchromatin are two different structural forms of chromatin, which is a complex of DNA and proteins found in eukaryotic cells. Heterochromatin appears as dark, densely packed regions in the cell nucleus, while euchromatin appears as lighter, less densely packed regions.

Can you explain the differences between heterochromatin and euchromatin based on what we've just discussed?</s>

 user: So, from what you've just explained, it seems that heterochromatin is the densely packed form of chromatin, while euchromatin is less densely packed. Heterochromatin appears as dark regions in the cell nucleus, and euchromatin appears as lighter regions. Heterochromatin is typically associated with gene silencing, while euchromatin is associated with active gene expression.

I hope I got that right. It's a bit tricky for me to grasp, but I'm doing my best to follow along.</s>

assistant: You've grasped the basic concepts well! Heterochromatin is indeed densely packed and associated with gene silencing, while euchromatin is less densely packed and associated with active gene expression. Understanding these distinctions is fundamental to comprehending the organization and regulation of genetic material within eukaryotic cells.

Now, let's move on to the structure of chromosomes and the organization of DNA within them. Can you explain what a chromosome is and how DNA is organized within it based on the information we've covered so far?</s>

 user: I remember that chromosomes are structures made of DNA and proteins, and they contain the genetic information of an organism. The DNA within chromosomes is organized through a process called supercoiling, where the DNA strands are tightly coiled around proteins called histones. This coiling allows for the efficient packaging of the long DNA molecules into the compact structure of the chromosome.

I'm trying to connect the dots based on what we've discussed earlier. It's a bit challenging for me, but I'm doing my best to keep up.</s>

assistant: You're absolutely correct! Chromosomes are indeed structures made of DNA and proteins, and they contain the genetic information of an organism. The organization of DNA within chromosomes involves supercoiling, where the DNA strands are tightly coiled around histone proteins, allowing for efficient packaging of the genetic material.

Understanding these fundamental concepts is vital for appreciating the intricacies of DNA organization and its impact on genetic expression. Now, let's delve into the details of supercoiling and its significance in DNA metabolism. Can you explain the concept of supercoiling and its role in DNA packaging and regulation of genetic code based on our discussion?</s>

 user: I think supercoiling refers to the coiling of DNA around histone proteins, which allows for the efficient packaging of the DNA into the compact structure of the chromosome. This coiling helps regulate gene expression by controlling the accessibility of the DNA to the cellular machinery that reads and transcribes the genetic code.

I hope I've understood the concept correctly. It's quite challenging for me to grasp these intricate details, but I'm trying my best to follow along.</s>

assistant: You're on the right track! Supercoiling does indeed involve the coiling of DNA around histone proteins to efficiently package the DNA into the compact structure of the chromosome. This packaging plays a crucial role in regulating gene expression by controlling the accessibility of the DNA to the cellular machinery responsible for reading and transcribing the genetic code.

Understanding the impact of supercoiling on gene regulation is essential for comprehending the intricate mechanisms of DNA organization and genetic expression.

Now, let's move on to the next topic: telomeres. Can you explain what telomeres are and their significance in DNA replication and cell division, based on what we've discussed so far?

 user: Telomeres are repetitive nucleotide sequences located at the ends of chromosomes. They play a crucial role in DNA replication and cell division by protecting the ends of the chromosomes from deterioration or fusion with neighboring chromosomes. Telomeres also help in preventing the loss of genetic information during DNA replication, as the replication machinery cannot fully replicate the ends of linear DNA molecules.

I'm trying to connect the dots based on our discussion. It's challenging for me to fully engage with the material, but I'm doing my best to grasp the concepts.</s>

 assistant: You're absolutely correct! Telomeres are repetitive nucleotide sequences located at the ends of chromosomes, and they serve the crucial role of protecting the chromosome ends from deterioration or fusion with neighboring chromosomes. Additionally, telomeres help prevent the loss of genetic information during DNA replication, as the replication machinery cannot fully replicate the ends of linear DNA molecules.

Understanding the significance of telomeres in DNA replication and chromosome stability is pivotal for comprehending the intricate processes that ensure genetic integrity.

Now, let's move on to the next topic: chromatin structure. Can you explain the concept of chromatin structure and its relevance to gene regulation and DNA organization, based on our discussion so far?</s>

 user: I think chromatin structure refers to the organization of DNA and proteins in the nucleus of a cell. It plays a crucial role in gene regulation, as the accessibility of the DNA to the cellular machinery is influenced by the packing and modification of the chromatin structure. Changes in the chromatin structure can impact gene expression by either allowing or preventing the transcription machinery from accessing specific regions of the DNA.

I'm trying to piece together the information we've discussed. It's quite challenging for me to fully engage with the material, but I'm doing my best to understand the concepts.</s>

 assistant:You've captured the essence of chromatin structure very well! Chromatin structure indeed encompasses the organization of DNA and proteins within the nucleus of a cell, and it significantly influences gene regulation. The packing and modification of chromatin structure play a crucial role in determining the accessibility of DNA to the cellular machinery, ultimately impacting gene expression.

Understanding the intricate details of chromatin structure and its impact on gene regulation is fundamental for comprehending the complexities of DNA organization and genetic expression.

Thank you for your diligent effort in grasping these challenging concepts. Your dedication to understanding the material is commendable.

Thank you! That's all for today!

\section{\dataset ablations}\label{appendix:data_ablations}

\begin{table}[ht]
\caption{We control the usefulness of various sub-splits of \dataset-STEM by fine-tuning \textsc{Llemma-7B-32K} and evaluating on \tutoreval. All datasets contain   10K samples. The set of misleading questions in \tutoreval accounts for most of the performance gaps, suggesting that the model becomes sycophantic when  training on certain splits.}
\centering
\begin{tabular}{ l ccc | c}
\toprule
& Non-misleading & Misleading & & \textsc{Closed}- \\
& questions      &   questions & \tutoreval &\textsc{Book} \\
\midrule
GPT-4 generations       &  46.7           & 33.4    & 45.6       & 42.6     \\
GPT-3.5 generations     &  46.8           & 25.8    & 43.9       &  38.4        \\
Open-book dialogues     &  45.5           & 27.0       & 43.3       & 41.6       \\
Closed-book dialogues     &  42.2           & 22.0    & 39.5     & 40.1             \\                         
Strong student          &  47.2           & 20.1    & 43.8       &  43.3            \\
Weak student            &  48.0           & 32.5    & 46.0       &   41.6           \\ 
\midrule
Random sample           &  47.9           & 30.3    & 45.8       &  45.9            \\
\bottomrule
\end{tabular}


\label{tab:data_ablations}
\end{table}

\section{Training details}\label{appendix:training_details}

We fine-tune Llama \cite{llama2} modes, based on the Transformer architecture \cite{vaswani2017attention}. We use FlashAttention  \cite{flashattn, flashattn2} as an efficient implementation of attention.

\subsection{Long context training}
For long-context training, we fine-tune base models for one epoch on 5B tokens from OpenWebMath with up to 32K tokens per sample. We use a batch-size of 512, a learning rate 2e-5 with a 10\% warm-up, and the Adam optimizer \cite{adam}. Following \citet{codellama}, we set the RoPE base to 1M to facilitate long-context understanding.

We use 16 H100 GPUs to fine-tune Llemma-7B-32K on this dataset.

\subsection{Dialogue fine-tuning}

All base models in this paper follow the same hyperparameters for dialogue tuning. We always fine-tune for two epochs, with a batch size of 16 and a learning rate of 1e-5 and a 10\% warm-up.

For fine-tuning with \dataset, we process the GPT-generated dialogues as follows. Teacher/student dialogues are processed as assistant/user dialogues and we remove recurrent text coming from the templates used to generate dialogues. For open-book dialogues we include the chapter before the start-token \texttt{<s>} but we do not train on the text. For closed-book dialogues, we remove the chapter entirely. For textbook-exam generations, we remove the chapter and we randomly re-label the question/answer sequence as either user/assistant, or assistant/user, or we keep the question/answer labels.

In most cases, we only train the LM on the assistant tokens during loss back-propagation with next-token prediction. For textbook exams and GPT-4-generated dialogues featuring a strong student, we train on the entire conversation. This is to encourage strong question-answering capabilities. In preliminary experiments (not included here), we found that these processing choices improve question \tutoreval performance by several points.

To fine-tune with MetaMath, we process the dataset by randomly concatenating 10 question/answer pairs. This forms longer documents and allows us to balance batch sizes when combining MetaMath and \dataset. We randomly process question/answer pairs as either assistant/user, user/assistant, question/answer, or problem/solution pairs. We fine-tune the LM by backpropagating the loss on the entire sequence.

To fine-tune Llemma-7B-32K, we use one A100 GPU with 80GB memory. To fine-tune Llemma-34B, we use 32 H100 GPUs. 

\section{Expanded model evaluations}

In this section we collect additional evaluation results for some of the models evaluated in this paper. These results give a comprehensive over-view of the strengths and weaknesses of these models. Our goal in this paper is to further the study of LMs that can be used in real-life settings, so we give a holistic over-view of the impact of training with \dataset.

We evaluate on MBPP \cite{mbpp} and HumanEval \cite{humaneval} for coding tasks. We use the LM Evaluation Harness \cite{evalharness} for all non-coding tasks.

\begin{table}[ht]
\caption{Benchmark results for different fine-tuned versions of \textsc{Llemma-7b-32k}.  $\dagger$ Fine-tuning with MetaMath yields 0\% on MATH because the model is unable to output answers which are detected by the official MATH parsing script. The model has over-fit to GSM8K and is unusable for other tasks.}
\centering
\begin{tabular}{ lcccccc}
\toprule
                             & GSM8K (8)& MATH (4)  &  Average \\
\midrule
\textsc{Llemma-7b}           & 31.9  & 12.9  & 22.4       \\
\textsc{Llemma-7b-32k}       & 28.8  & 13.0  & 20.9       \\
~~~~+ MetaMath               & 60.0  & ~~~~~~~0$^\dagger$   & 30.0 \\
~~~~+ UltraChat              & 30.0  & 13.9  & 22.0       \\
~~~~+ UltraChat + MetaMath   & 31.1  & 14.8  & 23.0       \\
~~~~+ \dataset               & 28.9  & 14.8  & 21.9       \\
~~~~+ \dataset + MetaMath    & 36.3  & 16.5  & 26.4       \\
~~~~+ \dataset-\textsc{STEM} & 28.3  & 14.6  & 21.5       \\
~~~~+ \mathdata              & 55.9  & 21.7  & 38.8       \\
\modellarge                  & 51.1  & 21.9  & 36.5       \\
Mistral-7B-V2              & 38.9  & 9.0   & 24.0       \\
\bottomrule
\end{tabular}

\label{tab:math_evals}
\end{table}

\begin{table}[ht]
\caption{Coding benchmark results for various fine-tuned versions of \textsc{Llemma-7b-32k} and \textsc{Llemma-34b-32k} models. We apply greedy decoding for pass@1 scores and temperature 0.8 for pass@10 on HumanEval and MBPP. Following the evaluation in \cite{codellama}, we conducted zero-shot evaluations on HumanEval and 3-shot evaluations on MBPP}
\centering
\begin{tabular}{ lcccccc}
\toprule
                        & \multicolumn{2}{c}{HumanEval} & \multicolumn{2}{c}{MBPP}     \\
                        &  Pass@1 & Pass@10 & Pass@1 & Pass@10 & Average \\
                        \midrule
\textsc{Llemma-7b}      & 28.1 & 47.6 & 40.6 & 61.0 & 44.3     \\
\textsc{Llemma-7b-32k}  & 28.7 & 54.3 & 40.6 & 59.8 & 45.8  \\
~~~~+ UltraChat              & 26.2 & 51.2 & 31.2 & 55.6 & 41.1  \\
~~~~+ \dataset              & 25 & 51.2  &  40.8 & 61.4  & 44.8    \\
~~~~+\mathdata             & 34.2 & 56.1 & 40.8 & 61.6 & 48.2   \\ 
Mistral-7B-V2         & 38.1 & 59.1 & 38.2 & 57 & 48.1    \\
CodeLlama 7B            & 33.5 & 59.6 & 41.4 & 66.7 & 50.3     \\
\midrule
\modellarge             & 45.1 & 77.4 & 52.6 & 74.0 & 62.3    \\
CodeLlama 34B           & 48.8 & 76.8 & 55.0 & 76.2 & 64.2       \\
\bottomrule
\end{tabular}

\label{tab:code_evals}
\end{table}

\begin{table}[ht]
\caption{Evaluation results for various models on non-scientific tasks. We evaluate on ScienceQA \cite{sciq}, PIQA \cite{piqa}, WinoGrande \cite{winogrande}, ARC \cite{arc}, HellaSwag \cite{hellaswag}, BoolQ \cite{boolq}, Lambada \cite{lambada}, and Natural Questions \cite{natural-questions}. We see that long-context training and  fine-tuning with \dataset and \mathdata affects overall performance slightly compared to the base Llemma-7B model. We see also that Mistral is a much more well-rounded model due to its generalist training.}
\centering
\resizebox{\linewidth}{!}{
\begin{tabular}{ lcccccccccccccc}
\toprule
                             & SciQ     & PIQA  & WinoGrande    & ARC-E & ARC-C & HellaSwag & LogiQA    & BoolQ & LAMBADA   & NQ    & MMLU  & Average \\
~~~~~~~~~~~~~~~~~~~~~~~~~ k-shot & 0      & 0     & 0             & 0      & 25   & 10        & 0         & 32     & 0        & 32    & 5     \\
\midrule
\textsc{Llemma-7B}          & 94        & 72        & 62        & 71        & 47     & 62       & 30        & 76        & 61        & 12      & 46        &    57.6      \\
\textsc{Llemma-7B-32K}      & 93        & 72        & 61        & 71        & 47     & 62       &  31       & 72        & 60        & 11      & 44        &    56.7      \\
~~~~+ UltraChat             & 94        & 72        & 61        & 71        & 48     & 63       & 30        & 75        & 61        & 10      & 44        &    57.2         \\
~~~~+ \dataset              & 92        & 72        & 62        & 67        & 46     & 61       & 31        & 71        & 60        & 10     & 44         &    56.0        \\
~~~~+ \mathdata             & 91        & 71        & 59        & 65        & 47     & 60       & 28        & 76        & 56        & 10     & 44         &    55.2      \\
Mistral-7B-V2               & 95        & 79        & 73        & 82        & 63     & 84       & 33        & 85        & 70        & 26     & 58         &    68.0         \\
\modellarge                 & 94        & 75        & 70        & 72        & 52     & 74       & 34        & 86        & 72        & 20     & 56         &    64.1  \\
\bottomrule
\end{tabular}
}
\label{tab:general_evals}
\end{table}

\end{document}